\documentclass[10pt,twocolumn,letterpaper]{article}

\usepackage{cvpr}
\usepackage{times}
\usepackage{textcomp}
\usepackage{epsfig}
\usepackage{graphicx}
\usepackage{amsmath}
\usepackage{amssymb}
\usepackage{hhline}
\usepackage{enumitem}
\usepackage{multirow}
\newenvironment{packed_enum}{
\begin{itemize}[leftmargin=*]
  \setlength{\itemsep}{0.6pt}
  \setlength{\parskip}{0pt}
  \setlength{\parsep}{0pt}
}{\end{itemize}}
\usepackage{lipsum} 
\usepackage{tabularx,colortbl}

\newcolumntype{P}[1]{>{\centering\arraybackslash}p{#1}}
\newcolumntype{C}[1]{>{\centering\arraybackslash}c{#1}}

\newcommand{\WStwo}[2] {IG-#1-#2}
\newcommand{\WSone}[1] {IG-#1}
\usepackage[pagebackref=true,breaklinks=true,letterpaper=true,colorlinks,bookmarks=false]{hyperref}

\cvprfinalcopy 

\def\cvprPaperID{5874} 

\ifcvprfinal\fi
\begin{document}

\title{Large-scale weakly-supervised pre-training for video action recognition}

\author{Deepti Ghadiyaram, Matt Feiszli, Du Tran, Xueting Yan, Heng Wang, Dhruv Mahajan
\\
Facebook AI\\
{\tt\small \{deeptigp, mdf, trandu, xyan18, hengwang, dhruvm\}@fb.com}
}

\maketitle

\begin{abstract}
Current fully-supervised video datasets consist of only a few hundred thousand videos and fewer than a thousand domain-specific labels. This hinders the progress towards advanced video architectures. This paper presents an in-depth study of using large volumes of web videos for pre-training video models for the task of action recognition. Our primary empirical finding is that pre-training at a very large scale (over $65$ million videos), despite on noisy social-media videos and hashtags, substantially improves the state-of-the-art on three challenging public action recognition datasets. Further, we examine three questions in the construction of weakly-supervised video action datasets. First, given that actions involve interactions with objects, how should one construct a \emph{verb-object} pre-training label space to benefit transfer learning the most? Second, frame-based models perform quite well on action recognition; is pre-training for good image features sufficient or is pre-training for spatio-temporal features valuable for optimal transfer learning? Finally, actions are generally less well-localized in long videos vs. short videos; since action labels are provided at a video level, how should one choose video clips for best performance, given some fixed budget of number or minutes of videos? 

\end{abstract}

\vspace{-0.15in}
\section{Introduction}
It is well-established \cite{decaf, imagenetTransfer} that pre-training on large datasets followed by fine-tuning on target datasets boosts performance, especially when target datasets are small \cite{posetrack, poseEstimation, segment, oflow}. Given the well-known complexities in constructing large-scale fully-supervised video datasets, it is intuitive that large-scale \textit{weakly-supervised} pre-training is vital for video tasks.

Recent studies \cite{laurens, uruImage, jft} have clearly demonstrated that pre-training on hundreds of millions (billions) of noisy web images significantly boosts the state-of-the-art in object classification. While one would certainly hope that successes would carry over from images \cite{laurens, uruImage, jft} to videos, action recognition from videos presents certain unique challenges that are absent from the image tasks.

First, while web images primarily face the challenge of \textit{label noise} (i.e., missing or incorrect object labels), for videos in the wild, the challenges are two-fold: label noise and \textit{temporal noise} due to the lack of localization of action labels. In real-world videos, a given action typically occupies only a very small portion of a video. In stark contrast, a typical web image is a particular moment in time, carefully selected by its creator for maximal relevance and salience.

Second, in prior work on images, labels were restricted to scenes and objects (i.e., nouns). However, action labels (eg: ``catching a fish'') are more complex, typically involving at least one verb-object pair. Further, even at large scale, many valid verb-object pairs may be observed rarely or never at all; for example, ``catching a bagel'' is an entirely plausible action, but rarely observed. Therefore, it is natural to inquire: is it more useful to pre-train on labels chosen from marginal distributions of nouns and verbs, do we need to pre-train on the observed portion of the joint distribution of (verb, noun) labels, or do we need to focus entirely on the target dataset's labels? How many such labels are sufficient for effective pre-training and how diverse should they be?

Third, the temporal dimension raises several interesting questions. By analogy to images, short videos should be better temporally-localized than longer videos; we investigate this assumption and also ask how localization affects pre-training. In addition, longer videos contain more frames, but short videos presumably contain more relevant frames; what is the best choice of video lengths when constructing a pre-training dataset?

Finally, we question whether pre-training on videos (vs images) is even necessary. Both frame-based models and image-based pre-training methods like inflation \cite{i3d} have been successful in action recognition. Is pre-training on video clips actually worth the increased compute, or, are strong image features sufficient?

In this work, we address all these questions in great detail. Our key aim is to improve the learned video feature representations by focusing exclusively on training data, which is complementary to model-architecture design. Specifically, we leverage over $65$ million public, user-generated videos from a social media website and use the associated hashtags as labels for pre-training. The label noise and temporal noise makes our training framework \textit{weakly-supervised}. Unlike \underline{all} existing fully-supervised video datasets \cite{thumos, hmdb, kth, hollywood, ucf, kinetics} which required expensive annotation, our training data is truly extensible to billion-scale without incurring any annotation costs. We effectively tackle the aforementioned challenges with label space and temporal noise, and demonstrate significant performance gains on various target tasks. Overall, we summarize our findings:
\begin{packed_enum}
\item {\textbf{Large-scale weak-supervision is extremely beneficial:}} We show that large-scale video data, despite not providing strong supervision, tremendously helps models of varied capacities in learning better features. Our experiments clearly demonstrate that content diversity and scale outdo label and temporal noise. 
\item \textbf{Impact of data volume and model capacity:} We report interesting findings on the effect of pre-training data size, data sampling strategies, model capacities, etc. For instance, we find that increasing the training data (Sec. \ref{sec:datapoints}) improves performance while increasing model capacity exhibits interesting behavior (Sec. \ref{sec:capacity}) .
\item \textbf{What is a suitable pre-training video label space?} We systematically construct pre-training label sets that vary in cardinality and type (e.g., verbs, nouns, etc.), and study their effects of target tasks (Sec. \ref{sec:labelEffect}). One key finding is that as in \cite{uruImage}, pre-training labels that overlap the most with the target labels improve performance.
\item \textbf{Short vs. long videos for pre-training?} We study the effect of pre-training on short vs. long videos (Sec. \ref{sec:vidLengths}) and show that (a) for a fixed \textit{video length budget} (e.g., 400K minutes of total training video duration), it is beneficial to choose a large number of short videos as they supply succinct localized actions compared to fewer long videos, (b) for a fixed \textit{video budget} (e.g., 5M
), choosing longer videos are beneficial as they offer diverse content. 
\item \textbf{Do we need pre-training on video data?} 
We investigate the true value of pre-training using video data. We show that it is necessary to use videos as opposed to video frames or images followed by inflation~\cite{i3d} to achieve better performance when operating at scale (Sec. \ref{sec:image2video}). 
\item {\textbf{State-of-the-art results:}} We achieve a top-1 accuracy of $\boldsymbol{81.3\%}$ on Kinetics, a $3.6\%$ boost over the previous state-of-the-art \cite{nonlocal} (Sec. \ref{sec:sota}). While the gains in \cite{nonlocal} were achieved via architectural innovation, increased compute, etc., our boost is purely from pre-training a simple architecture (R(2+1)D \cite{r2plus1D}) at scale. On EPIC Kitchens action recognition challenge \cite{epicKitchens}, we achieve an accuracy of $\boldsymbol{25.6\%}$ on unseen (S2) test data, an improvement of $4.6\%$ over the top entry in the leadership board at the time of submission. On Something-something \cite{something-something}, we achieve an accuracy of $\boldsymbol{51.6\%}$, a $2.1\%$ improvement over the previous state-of-the-art \cite{somethingeco}.
\end{packed_enum}
\section{Related Work} \label{sec:relatedWork}
\noindent \textbf{Learning from weak supervision: }Given the known challenges in gathering exhaustive annotations for various image and video tasks, leveraging object labels and other meta information to supply weak supervision \cite{deepDetect, segment1, paris, wildcat, segment3, segment2, segment4, signLang,  freeLocalization, segment5, segment6, visualRelations, segment7, weakVideo, shi2017transfer} is a widely-adopted approach. Orthogonal to this strategy, our work investigates transfer learning benefits when networks are pre-trained on weakly-supervised data, i.e., data afflicted with label and temporal noise. While novel changes to architectures have been proposed \cite{noise1, noise2} to combat label noise, our experiments demonstrate that large-scale training of an existing video architecture \cite{r2plus1D} makes it noise-resilient.\\
\noindent\textbf{Dataset sources:} Many prior approaches use Internet as a natural source of content \cite{bergamo2010exploiting, chen2015webly, neil, everything, farhadi2010every, fergus2010learning, f1, f2, f3, uruImage, im2text, schroff2011harvesting, annotate} and the associated search queries, hashtags, or user-comments as labels to construct datasets. Most large-scale video datasets \cite{youtube8m, yfcc100m, sports1m} were constructed by first curating a label taxonomy, analyzing the text metadata surrounding YouTube or Flickr videos, followed by some manual cleaning of the non-visual labels. \cite{v3, v4, v1, v2} analyzed movie scripts for automatic annotation of human actions for recognizing and localizing actions and actors. 
Our proposed work uses web-data and the associated text to supply weak supervision during pre-training.\\
\noindent \textbf{Pre-training on large datasets:} Since datasets for complex tasks such as object detection, segmentation and action recognition in videos are in smaller order of magnitude compared to ImageNet \cite{imagenet}, pre-training on larger, auxiliary data followed by fine-tuning on target tasks \cite{i3d, decaf, segment, laurens, l2, uruImage, jft, r2plus1D, nonlocal} is very popular. Indeed, \textit{inflation} \cite{i3d} was proposed to exclusively leverage ImageNet instantiation by way of converting $2D$ filters to $3D$, given that training $3D$ models is computationally expensive. In this work, we show that pre-training on video clips performs significantly better than pre-training on image/video frames followed by inflation (Sec. \ref{sec:image2video}).
\vspace{-0.07in}
\section{Weak-supervision of video models}
We harness millions of public videos from a social media website and use the associated hashtags as labels to supply weak supervisory signals while training video models. We construct and experiment with a variety of weakly-supervised datasets, which we describe next.
\subsection{Source datasets} \label{sec:igdataset}
To construct pre-training video datasets, we use several \textit{seed} action label sets and gather videos that are related to these labels. Specifically, for a given seed action label ``catching a fish," we first construct all possible meaningful phrases (i.e., relevant hashtags) from it by taking the original and stemmed versions of every word in the seed label and concatenating them in all possible permutations. As an example, $\mathtt{relevant\_hashtags}(``catching\ a\ fish") = $ $\{\#catchingafish,$ $\#catchfish,$ $\#fishcatching, ... \}$. We then download public videos that are tagged with at least one of the hashtags from the set of $\mathtt{relevant\_hashtags}(``catching\ a\ fish")$ and associate them with the initial seed label. We use the seed labels as the final labels for videos during pre-training. 
\vspace{-0.1in}
\subsubsection{Properties of the source datasets} \label{sec:datasetprop}
\noindent \textbf{Seed labels:} As we describe in Sec. \ref{sec:labelEffect}, we study the effect of pre-training on different types of labels by considering four different seed label sets. The resulting source datasets are summarized in Table \ref{tbl:sourceDatasets}. We use a notation $\mathtt{IG-source-size}$ throughout this paper, where \textit{source} indicates the seed label set used and \textit{size} indicates the number of videos\footnote{We use $\mathtt{IG-source}$ as notation whenever we refer to pre-training data source alone throughout this paper.}. Our primary training set uses $400$ action labels from Kinetics \cite{kinetics17} as seed labels, resulting in $\mathtt{\WSone{Kinetics}}$ dataset comprising $359$ labels\footnote{For the remaining $41$ labels, we could not find sufficient videos (i.e., at least $50$ per label) using the approach detailed in Sec. \ref{sec:igdataset}.}. We also consider (a) the $1428$ hashtags that match the $1000$ synsets from ImageNet-1K \cite{imagenet}, thereby constructing an $\mathtt{IG-Noun}$ dataset\footnote{We get $1428 (> 1000)$ total hashtags because multiple hashtags may map to the same synset.}, (b) the $438$ verbs from Kinetics and VerbNet \cite{verbnet}, thus an $\mathtt{IG-Verb}$ dataset, and (c) all possible concatenations of the $438$ verbs and the $1428$ nouns from the above two seed label sets. We identify over $10,653$ such meaningful combinations\footnote{Note that this is far fewer than $438 * 1428 =$\texttildelow$600k$, as we discarded those concatenations which are not associated with at least $50$ videos.}, thus constructing an $\mathtt{IG-Verb+Noun}$ dataset. More details on dataset construction are provided in the supplementary material. We want to reiterate that no manual annotation was involved in constructing these datasets implying that there is a large amount of noise in the data.\\
\textbf{Long tail distribution:} Distribution of videos in all our source datasets is heavily long-tailed. To mitigate its effect on model training, we adopt the square root sampling approach \cite{sqrtSampling} when constructing pre-training data in our experiments as it proved to be most beneficial for images \cite{uruImage}.\\
\textbf{Diversity:} Unlike all benchmark datasets which contain short videos of localized actions, video lengths in our source datasets range from $1$ to $60$ seconds. Thus, the labeled action can occur anywhere in a video, resulting in a large amount of temporal noise, an aspect we study in Sec. \ref{sec:temporal}.

Given that we cannot make our datasets or the exact hashtags used public (as was the case with \cite{jft, uruImage}), we acknowledge that it is not possible for other research groups to reproduce our results at this time. Despite this limitation, we hope that the reader finds value in our wide-range of findings on various aspects of large-scale video pre-training that we describe in Sec. \ref{sec:experiments}. 
\begin{table}
\scriptsize
\begin{center}
\begin{tabular}{|l|c|c|}
\hline
Pre-training dataset & Total \#Videos & \#Labels\\
\hline
{\WSone{Kinetics}} & $65M$ & $359$ \\
{\WSone{Noun}} & $19M$ & $1428$ \\
{\WSone{Verb}} & $19M$ & $438$\\
{\WSone{Verb+Noun}} & $19M$ & $10653$\\
\hline
\end{tabular}
\end{center}
\vspace{-0.09in}
\caption{\scriptsize{Statistics of the weakly-supervised datasets constructed for pre-training.}}
\vspace{-0.23in}
\label{tbl:sourceDatasets}
\end{table}
\subsection{Target datasets} \label{sec:targetdataset}
\noindent Next, we describe the target datasets used in experiments.\\
\textbf{Kinetics} \cite{kinetics}: Kinetics is a multi-class dataset with \texttildelow$246K$ training videos ($400$ human action labels). We report performance on the $20K$ validation videos.\\
\textbf{EPIC-Kitchens} \cite{epicKitchens}: EPIC-Kitchens is a multi-class egocentric dataset with \texttildelow$28K$ training videos associated with $352$ noun and $125$ verb classes. For our ablation studies, following~\cite{epic1} we construct a validation set of unseen kitchen environments. We evaluate our best pre-trained model on validation, standard seen (S1: $8047$ videos), and unseen (S2: $2929$ videos) kitchens test datasets.\\
\textbf{Something-Something-v1} \cite{something-something} is a multi-class dataset of \texttildelow$86K$ training videos and $174$ fine-grained actions. We report results on the $11,522$ validation set.\vspace*{0.1in}\\
\textbf{Video deduplication:} We devise a pipeline to deduplicate videos in the source datasets that may overlap with any from the target dataset. To err on the side of caution, we adopt an aggressive low-precision high-recall strategy and remove any potential duplicates (eg: we removed \texttildelow$29K$ videos from the \WSone{Kinetics }dataset). Details are provided in the supplementary material.
\subsection{Pre-training Setup} 
\noindent \textbf{Models:} R(2+1)D-d \cite{r2plus1D}\footnote{{Source code: \url{https://github.com/dutran/R2Plus1D}}} is the fundamental architecture used for pre-training, where $d$ denotes model depth = $\{18, 34, 101, 152\}$. As in \cite{residual}, we construct models of depth $>\ 34$ by replacing simple temporal blocks with bottleneck blocks for computational feasibility. We direct the reader to the supplementary material for details.\vspace*{0.05in}\\
\noindent \textbf{Loss Function:} Our pre-training datasets are multi-label since multiple hashtags may be associated with any given video. The authors of \cite{laurens,uruImage} have observed that per-label sigmoid outputs with logistic loss do not work well for noisy labels. Hence, we follow a simple strategy of randomly assigning one of the associated hashtags to each video thereby formulating a multi-class problem, and use softmax activations with cross-entropy loss.\vspace*{0.05in}\\
\noindent \textbf{Training Details: }
Video frames are down-sampled to a resolution of $128\times171$ and each video clip is generated by cropping a random patch of size $112\times112$ from a frame. Video clips of either $8$ or $32$ frames are used in our experiments, and temporal jittering is also applied to the input. Synchronous stochastic gradient descent (SGD) is used to train our models on $128$ GPUs across $16$ machines using caffe2 \cite{caffe2}. When $32$ frames per input video clip are considered, each GPU processes $6$ videos at a time (due to memory constraints), while $16$ videos are processed at a time when $8$ frames per video clip are considered. Batch normalization (BN) is applied to all convolutional layers and the statistics \cite{batchNorm} are computed on each GPU. All pre-training experiments process $490M$ videos in total. Learning rate is set following the linear scaling procedure proposed in \cite{imageNethour} with a warmup. An initial learning rate of $0.192$ is used which is divided by $2$ at equal steps such that the total number of learning rate reductions is $13$ over the course of training.
\begin{figure*}[t]
\begin{center}$
\begin{array}{ccc}
\includegraphics[width=0.275\linewidth]{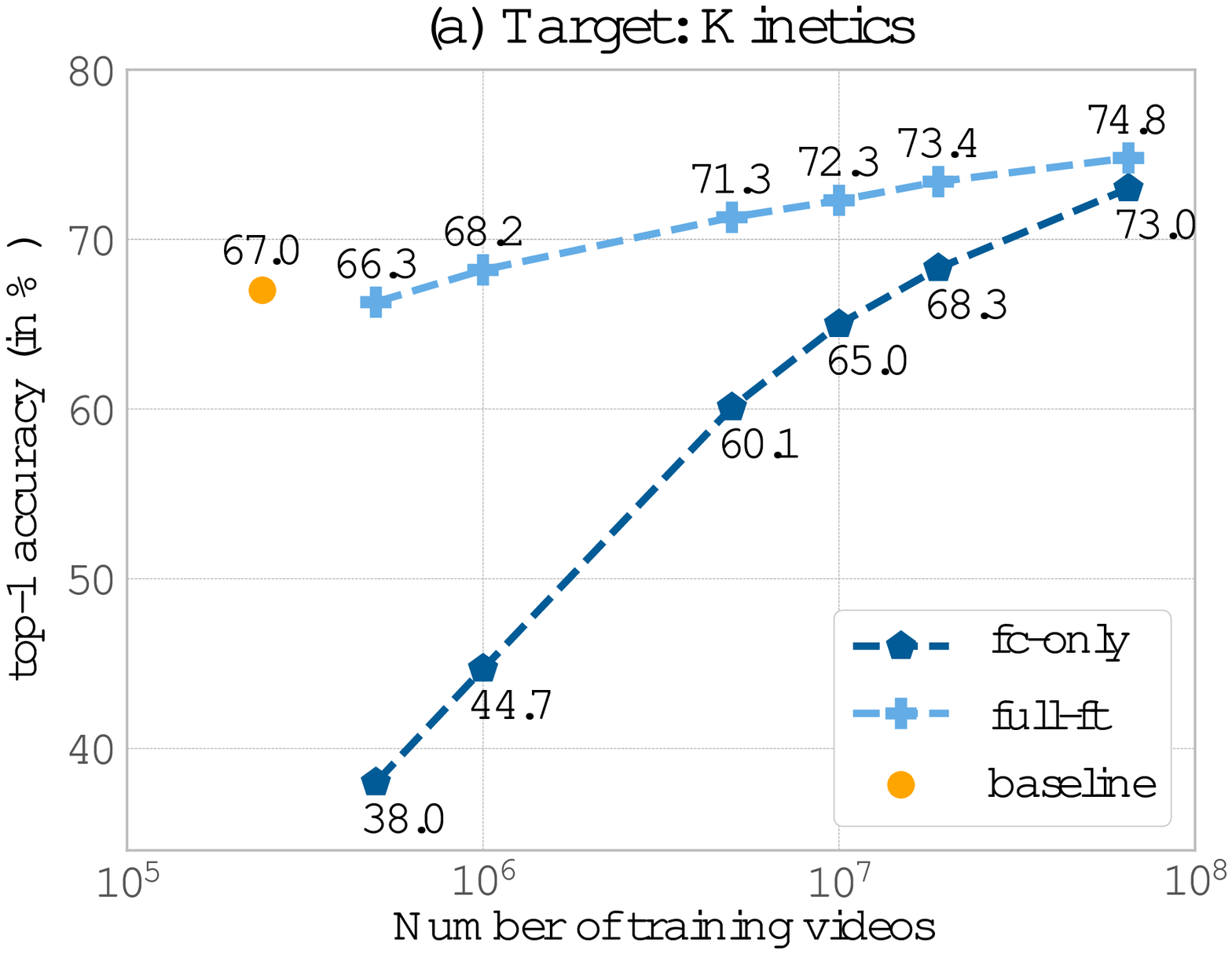} & 
\includegraphics[width=0.27\linewidth]{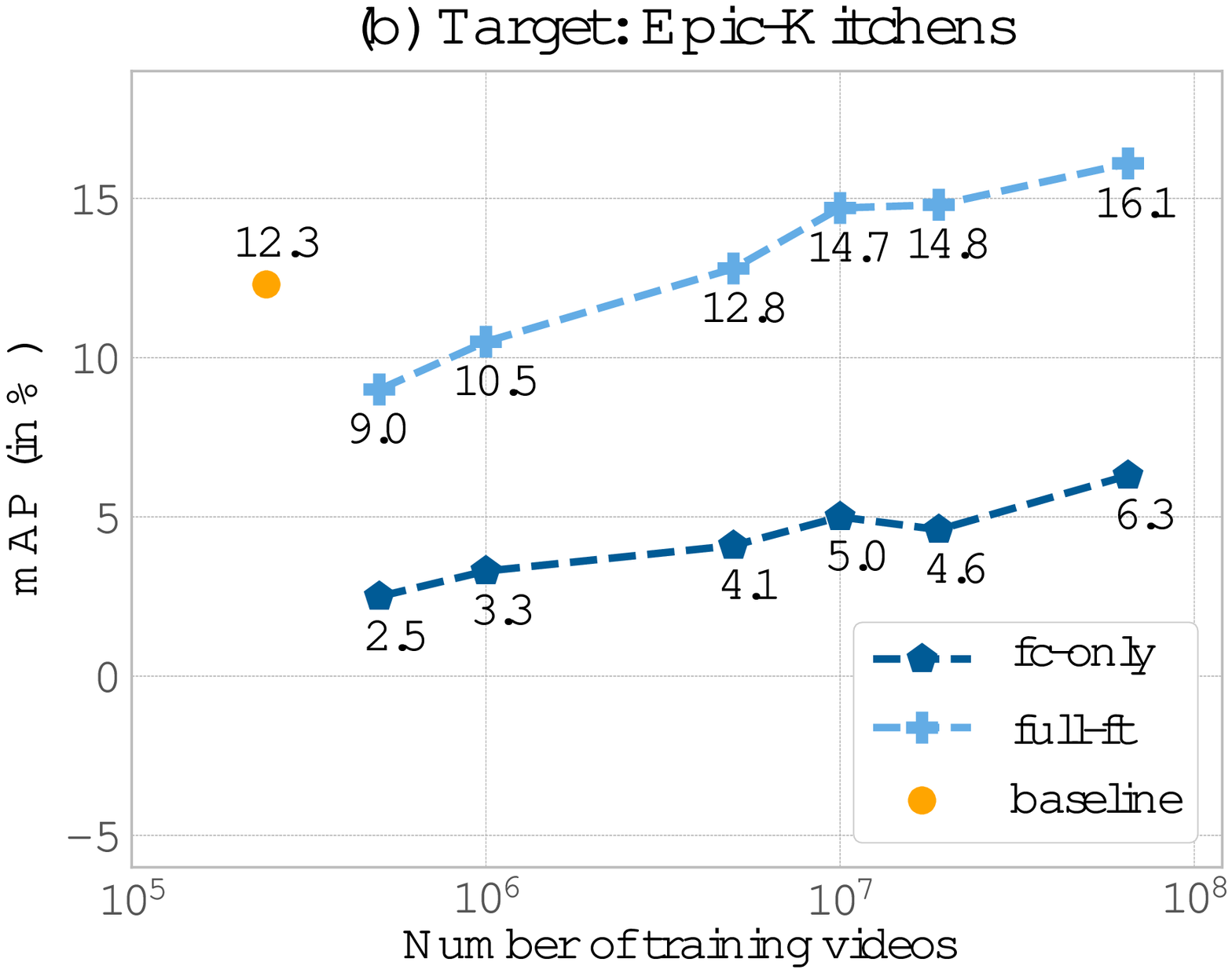} & 
\includegraphics[width=0.27\linewidth]{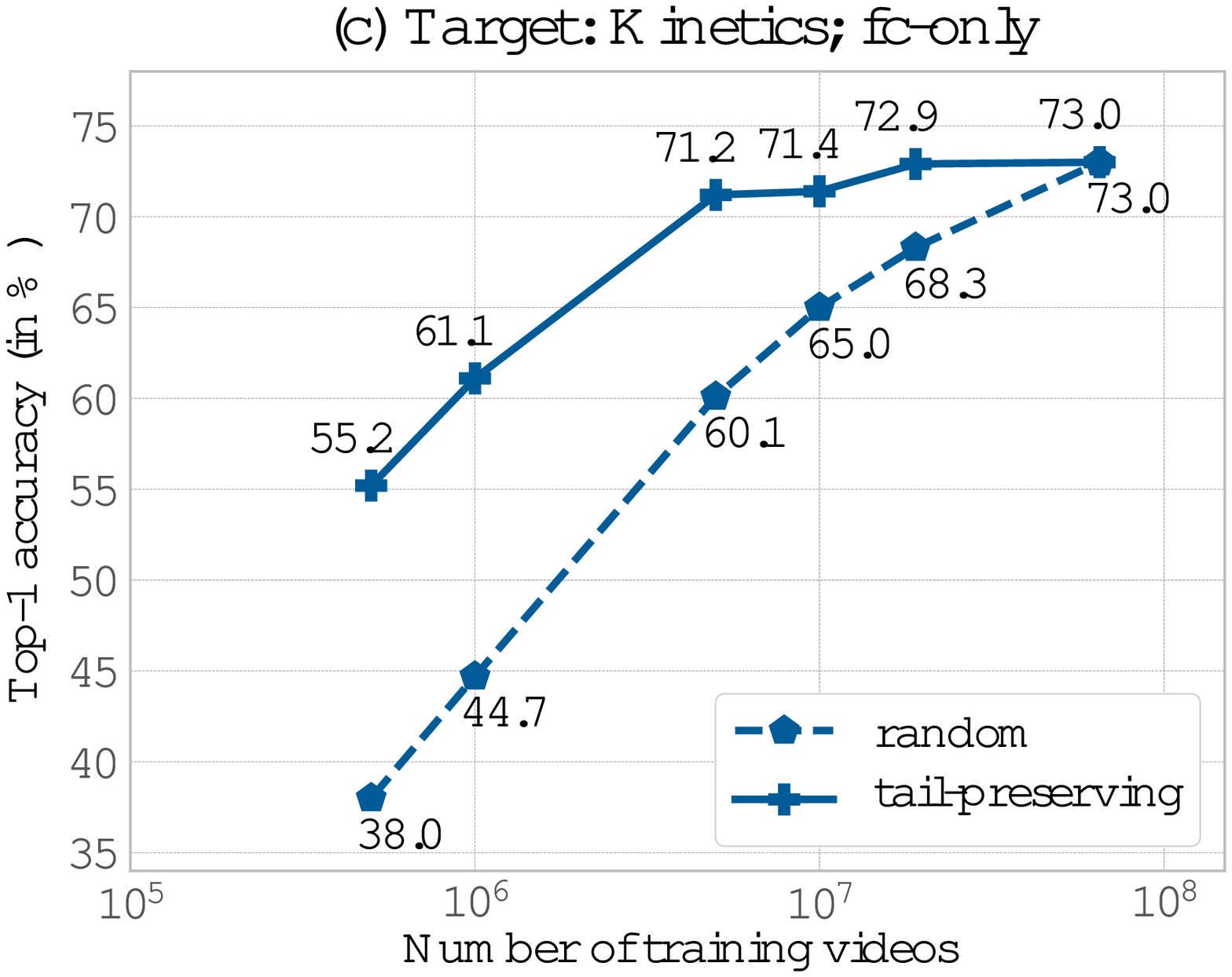} \\
\end{array}$ 
\caption{\scriptsize{Illustrating the effect of increasing the number of pre-training videos. For Kinetics, we train a R(2+1)D-34 model from scratch as baseline, while for EPIC-Kitchens, we pre-train R(2+1)D-34 on Kinetics as baseline (indicated in orange). Random sampling was used for experiments reported in (a) and (b). X-axis is in log-scale.}}
\vspace{-0.2in}
\label{fig:dataPoints}
\end{center}
\end{figure*}
\vspace{-0.05in}
\section{Experiments} \label{sec:experiments}
In this section, we study various aspects of large-scale weakly-supervised video pretraining. We first describe our evaluation setup and then report our extensive analysis on three aspects: (a) effect of scale, e.g., model capacity and pre-training data size, (b) design of the pre-training label space, and (c) temporal properties of videos. We also pre-train on benchmark datasets such as Sports-1M \cite{sports1m}, Kinetics \cite{kinetics} as competitive baselines.\vspace*{0.05in}\\
\noindent\textbf{Evaluation Setup: }As in \cite{uruImage}, we consider two scenarios:
\begin{packed_enum}
\vspace{-0.04in}
\item \textbf{Full-finetuning (full-ft) approach} involves bootstrapping with a pre-trained model's weights and training end-to-end on a target dataset. We do a grid search for best the hyper-parameters (learning rate etc.) on validation data constructed by randomly holding out ($10\%$) of training data. The hyper-parameters used for each experiment and target dataset are in the supplementary material. Full-ft approach has the disadvantage that it can potentially mask the absolute effect of pre-training for large target datasets.
\item \textbf{Fully-connected (fc-only) approach} involves extracting features from the final fc layer of a pre-trained model and training a logistic regressor on each target dataset. This approach evaluates the strength of the learned features without changing the network parameters.
\vspace{-0.1in}
\end{packed_enum}
For multi-class target datasets our loss function is a $L2$-regularized logistic regressor and we report \textit{accuracy}. For multi-label datasets, we use a per-label sigmoid output followed by logistic loss and report \textit{mAP}. During testing, center crops of $10$ clips uniformly sampled from each test video are considered, and the average of these $10$ clip predictions are used to obtain the final video-level prediction. 
\subsection{Effect of large-scale}
\subsubsection{Amount of pre-training data} \label{sec:datapoints}
To understand this question, we pre-train on different amounts of training data by constructing different data subsets - \WSone{Kinetics}-
$\{ 500K, 1M, 5M, 10M, 19M, 65M \}$.
R(2+1)D-34 models are independently trained on these data subsets on the exact same labels, with an input of $8$-frames per video and evaluated on Kinetics (Fig. \ref{fig:dataPoints} (a)) and EPIC-Kitchens (Fig. \ref{fig:dataPoints} (b)). 

As in \cite{uruImage, jft}, we observe that performance improves log-linearly with training data size indicating that more pre-training data leads to better feature representations. For Kinetics, with full-ft approach, pre-training using $65M$ videos gives a significant boost of $\boldsymbol{7.8\%}$ compared to training from scratch ($74.8\%$ vs. $67.0\%$). With increase in training data, performance gains are even more impressive when using fc-only approach, which achieves an accuracy of $73.0\%$ with $65M$ training videos, thus closely matching the accuracy from full-ft approach ($74.8\%$). On EPIC-Kitchens, using \WStwo{Kinetics}{65M} yields an improvement of $\boldsymbol{3.8\%}$ compared to using Kinetics for pre-training ($16.1\%$ vs. $12.3\%$). Compared with Kinetics, on EPIC-Kitchens, there is a larger gap in the performance between full-ft and fc-only approaches. This may be due to a significant domain difference in the pre-training and target label space. 

These plots indicate that despite the dual challenge of label and temporal noise, pre-training using millions of web videos exhibit excellent transfer learning performance.\vspace*{0.05in}\\
\noindent\textbf{Data Sampling:} Web data typically follows a Zipfian (long tail) distribution. When using only a subset of such data for pre-training, a natural question to ask is, if there are better ways to choose a data subset beyond random sampling. We design one such approach where we retain all videos from tail classes and only sub-sample head classes. We refer to this scheme as \textit{tail-preserving} sampling.

Figure~\ref{fig:dataPoints} (c) compares random and tail-preserving sampling strategies for Kinetics and reports performance obtained via fc-only approach. We observe that the tail-preserving strategy does consistently better and in fact, the performance saturates around $10M-19M$ data points. Hence, for all future experiments, we adopted tail-preserving sampling strategy when needed.
\vspace{-0.14in}
\subsubsection{Effect of model capacity} \label{sec:capacity}
Table~\ref{tbl:depth} reports the capacity of different video models and their effect on transfer learning performance. Specifically, we use \WStwo{Kinetics}{65M} to pre-train $4$ different R(2+1)D-d models, where $d = \{18, 34, 101, 152\}$ with input clip-length $32$. On Kinetics, we observe that increasing model capacity improves the overall performance by $3.9\%$. In comparison, when training from scratch, the accuracy improves only by $2.7\%$. Interestingly, on EPIC-Kitchens, pre-training either using \WStwo{Kinetics}{65M} or Kinetics (referred to as baseline) yield similar gains with the increase in model capacity. Unlike in~\cite{uruImage} where the transfer learning performance was observed to be bottlenecked by capacity, we see a saturation in performance when going from $d=101$ to $d=152$\footnote{For EPIC-Kitchens, we even observe a performance drop.}. Given that R(2+1)D-152 has higher GFLOPS compared to the largest image model in \cite{uruImage}, we believe that our model may be bottlenecked by the amount of pre-training data. Thus, using more than $65M$ training videos may further boost the accuracy. Additionally, inability to do long-range temporal reasoning beyond $32$ frames (due to memory constraints) may also be leading to this behavior. These questions are interesting to explore in the future.
\begin{table}[t]
\vspace{-0.1in}
\scriptsize
\setlength\extrarowheight{1.0pt}
\centering
\begin{center}
\begin{tabular}{|P{1.35cm}|P{0.78cm}|P{0.89cm}|P{0.58cm}|P{0.6cm}|P{0.58cm}|P{0.6cm}|}
\hline
& & & \multicolumn{2}{|c|}{Kinetics} & \multicolumn{2}{|c|}{Epic-Kitchens} \\
\hline
 Models & GFLOPS & \# params & full-ft & baseline & full-ft & baseline \\
\hline
R(2+1)D-18 & 83 & 33M & 76.0 & 69.3 & 20.8 & 14.8 \\
\hline
R(2+1)D-34 & 152 & 64M & 78.2 & 69.6 & 22.4	& 15.2 \\
\hline
R(2+1)D-101 & 176 & 86M & 79.1 & 71.7 & 24.9	& 17.1 \\
\hline
R(2+1)D-152 & 252 & 118M & 79.9 & 72.0 & 23.7	& 17.8 \\
\hline
\end{tabular}
\end{center}
\vspace{-0.1in}
\caption{\scriptsize{Performance when pre-trained models of varied capacities are fully-finetuned on Kinetics (top-1 accuracy) and Epic-Kitchens (mAP). For EPIC-Kitchens, as a baseline, we use a model pre-trained on Kinetics.
}} \label{tbl:depth}
\vspace{-0.2in}
\end{table}

\vspace{-0.18in}
\subsection{Exploring the pre-training label space} \label{sec:labelEffect}
Web videos and the associated (noisy) hashtags are available in abundance; hence it is natural to question: what constitutes a valuable pre-training label space for achieving superior transfer learning performance and how to construct one? Since hashtags are generally composed of nouns, verbs, or their combinations, and vary greatly in their frequency of occurrence, it is important to understand the trade-offs of different pre-training label properties (eg: cardinality and type) on transfer learning. In this section, we study these aspects in great detail.
\begin{figure*}[t]
\begin{center}$
\begin{array}{cccc}
\includegraphics[width=0.2\linewidth]{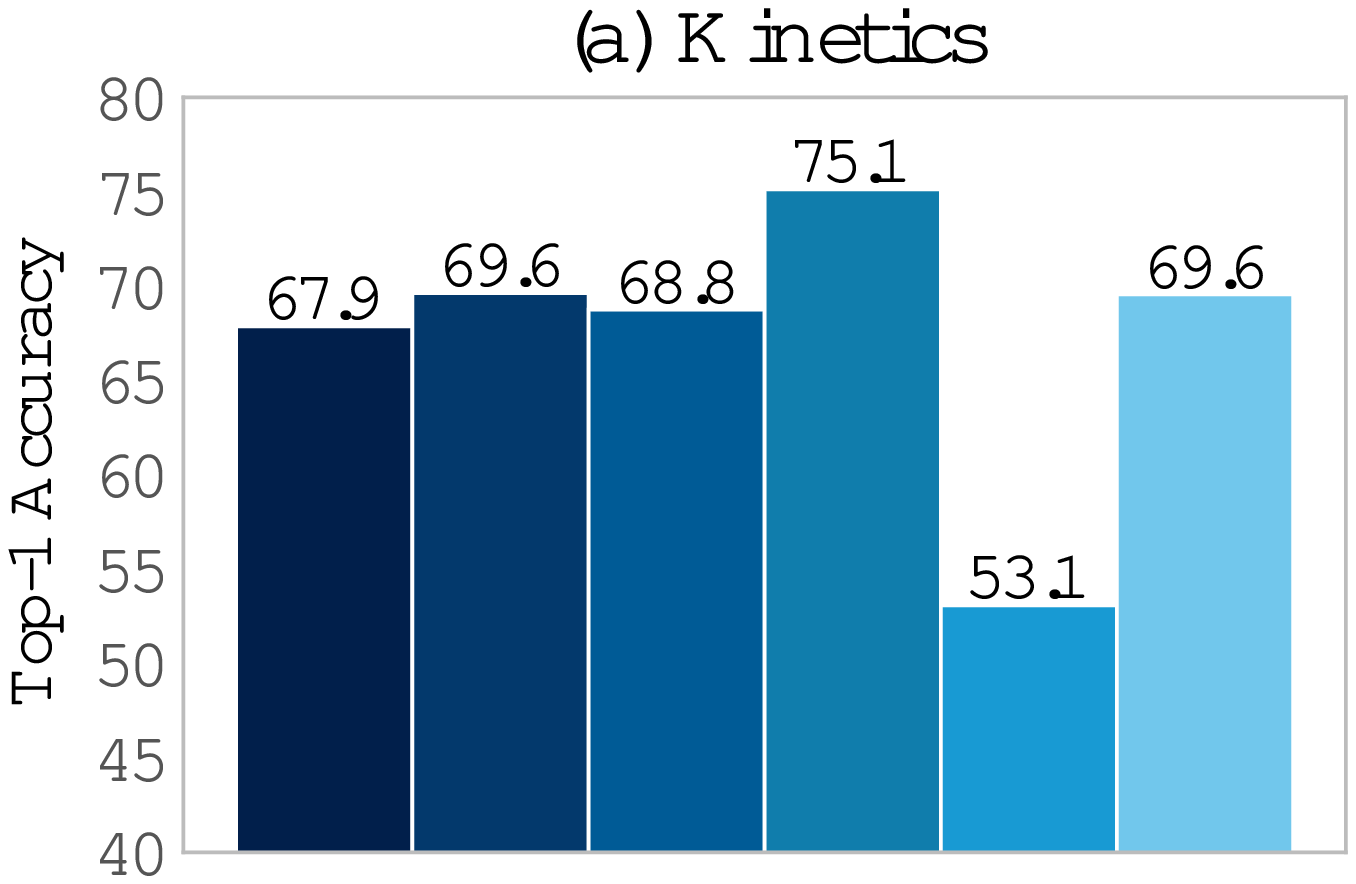} & 
\includegraphics[width=0.2\linewidth]{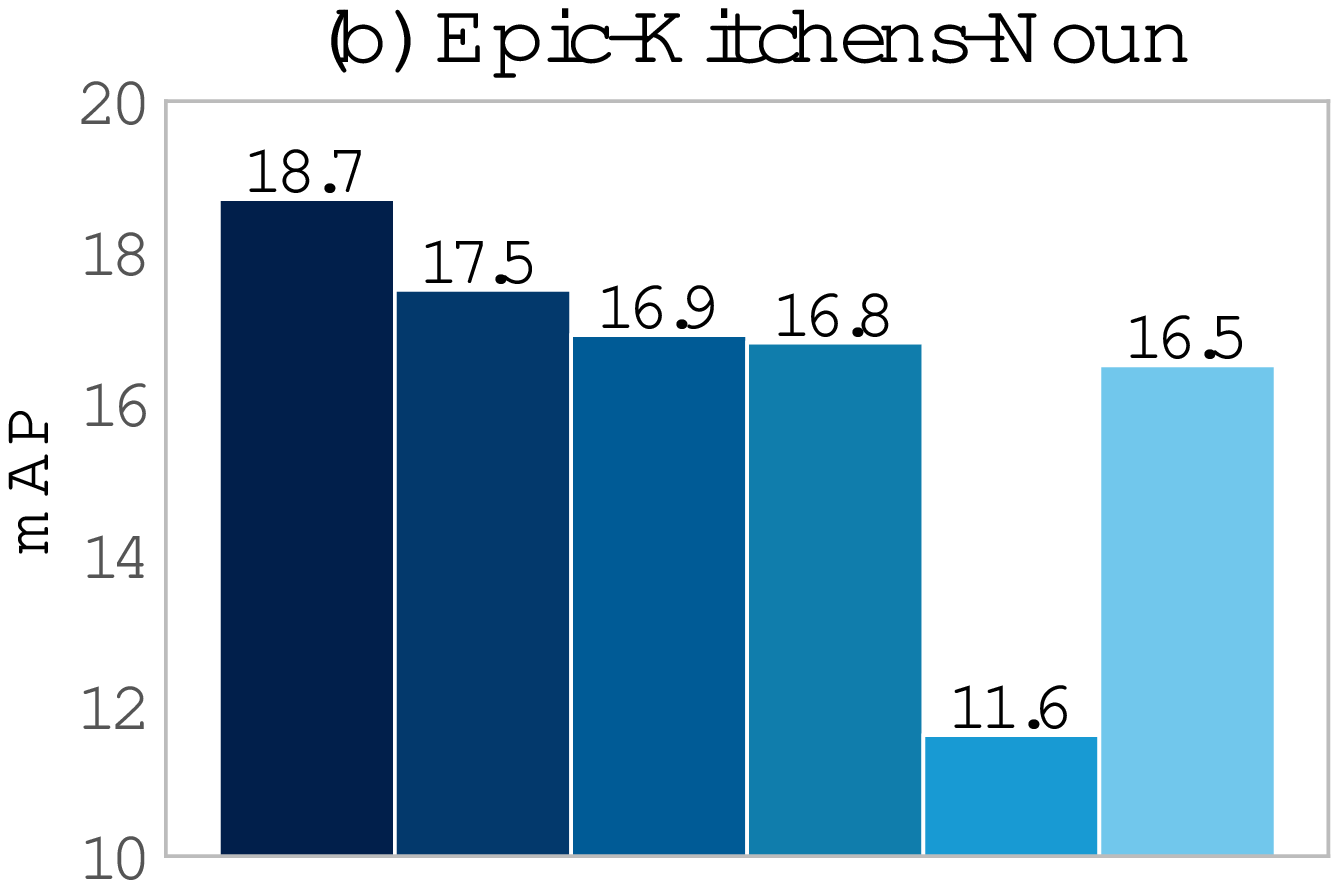} &
\includegraphics[width=0.2\linewidth]{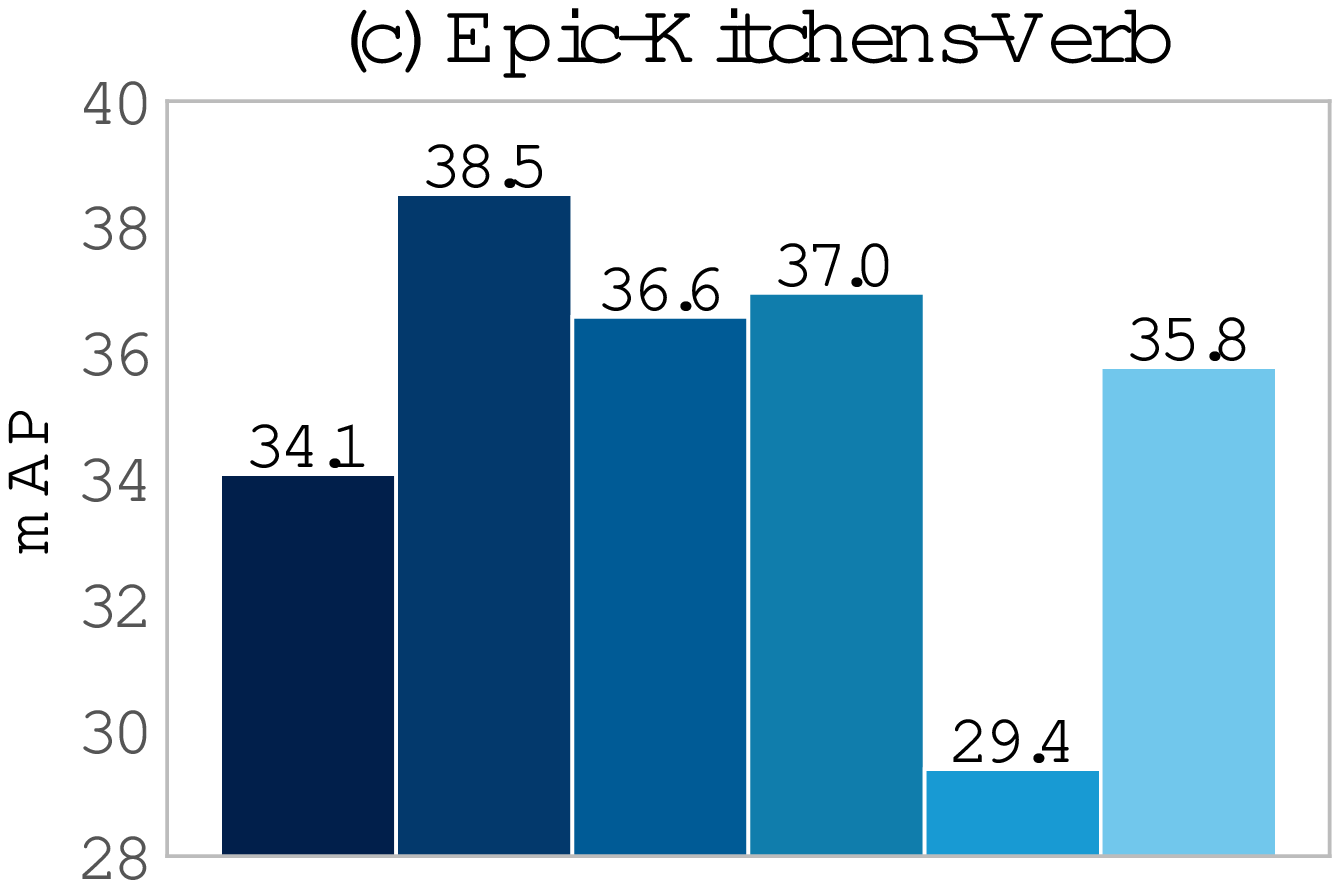} & 
\includegraphics[width=0.3\linewidth]{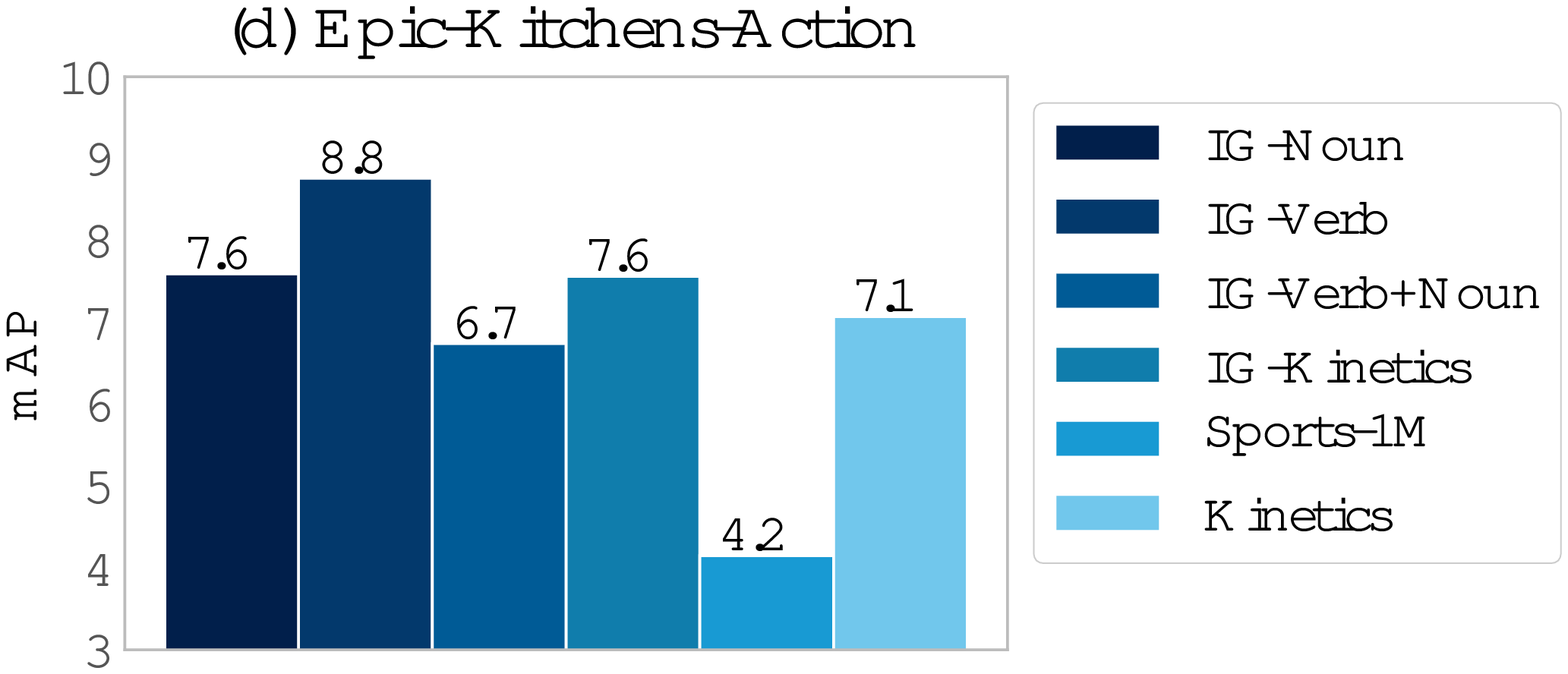} \\
\end{array}$ 
\caption{\scriptsize{(a) Top-1 accuracy on Kinetics and (b)-(d) mAP on the three Epic-Kitchens tasks after fc-only finetuning, when different source label sets are used (indicated in the legend). The results indicate that target tasks benefit the most when their labels overlap with the source hashtags. Best viewed in color.}}
\vspace{-0.28in}
\label{fig:labelType}
\end{center}
\end{figure*}
\vspace{-0.16in}
\subsubsection{Effect of the nature of pre-training labels}\label{sec:labelType} 
\vspace{-0.07in}
To study the type of pre-training labels that would help target tasks the most, as mentioned in Sec. \ref{sec:igdataset}, we systematically construct label sets that are verbs, nouns, and their combinations. Specifically, we use \WStwo{Kinetics}{19M}, \WStwo{Verb}{19M}, \WStwo{Noun}{19M}, and \WStwo{Verb+Noun}{19M} as our pre-training datasets. We use R(2+1)D-34 with clip-length of $32$ for training. From Fig. \ref{fig:labelType}, we may observe that for each target dataset, the source dataset whose labels overlap the most with it yield maximum performance. For instance, for Kinetics we see an improvement of \textit{at least} $5.5\%$, when we use \WStwo{Kinetics}{19M} for pre-training, compared to other pre-training datasets (Fig. \ref{fig:labelType}(a)). Pre-training on \WSone{Noun} benefits the noun prediction task of EPIC-Kitchens the most while \WSone{Verb} significantly helps the verb prediction task (at least 1.2\% in both cases, Fig. \ref{fig:labelType}(b) and (c)). We found an overlap of $62\%$ between \WSone{Verb} and the verb labels and $42\%$ between \WSone{Noun} and the noun labels in EPIC-Kitchens. Pre-training on Sports-1M performs poorly across all target tasks, presumably due to its domain-specific labels.
\begin{figure}[t]
\begin{center}
\includegraphics[width=0.45\linewidth]{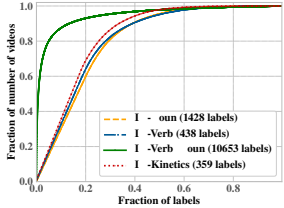}
\caption{\scriptsize{Cumulative distribution of the number of videos per label for the $4$ pre-training datasets discussed in Sec. \ref{sec:labelType}. The x-axis is normalized by the total number of labels for each dataset.}}
\label{fig:labelDist}
\vspace{-0.15in}
\end{center}
\end{figure}

Given that actions in EPIC-Kitchens are defined as verb-noun pairs, it is reasonable to expect that \WSone{Verb+Noun} is the most well-suited pre-training label space for EPIC-Kitchens-actions task. Interestingly, we found that this was not the case (Fig. \ref{fig:labelType} (d)). To investigate this further, we plot the cumulative distributions of the number of videos per label for all four pre-training datasets in Fig.~\ref{fig:labelDist}. We observe that though \WSone{Verb+Noun} captures all plausible verb-noun combinations leading to a very large label space, it is also heavily skewed (and hence sparse) compared to other datasets. This skewness in the \WSone{Verb+Noun} label space is perhaps offsetting its richness and diversity as well as the extent of its overlap with the EPIC-Kitchens action labels. Thus, for achieving maximum performance gains, it may be more effective to choose those pre-training labels that most overlap with the target label space while making sure that label distribution does not become too skewed. Understanding and exploiting the right trade-offs between these two factors is an interesting future research direction. 
\vspace{-0.29in}
\subsubsection{Effect of the number of pre-training labels}\label{sec:numLabels}
\vspace{-0.06in}
\begin{figure}[t]
\begin{center}$
\begin{array}{cc}
\includegraphics[width=0.4\linewidth]{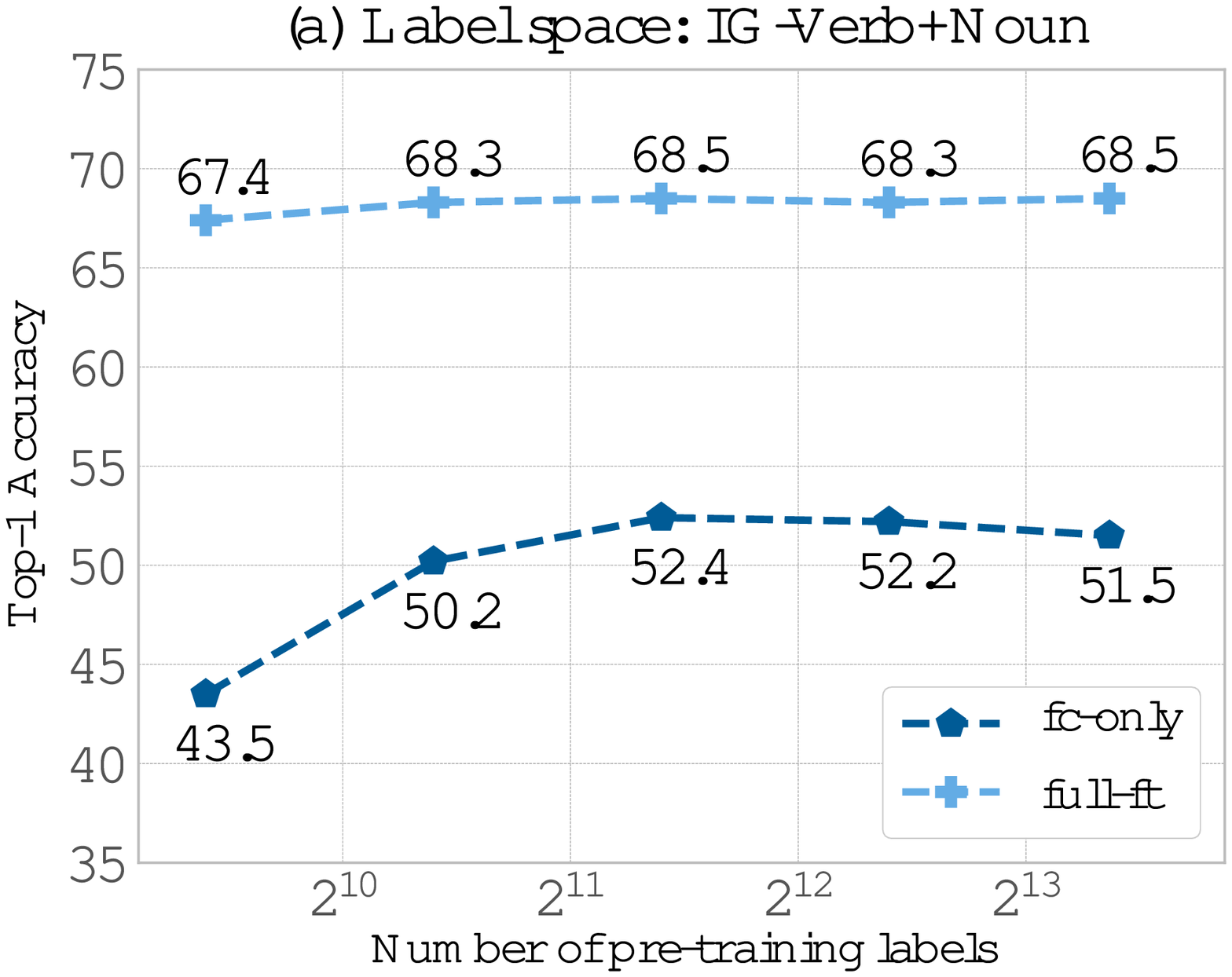} & 
\includegraphics[width=0.4\linewidth]{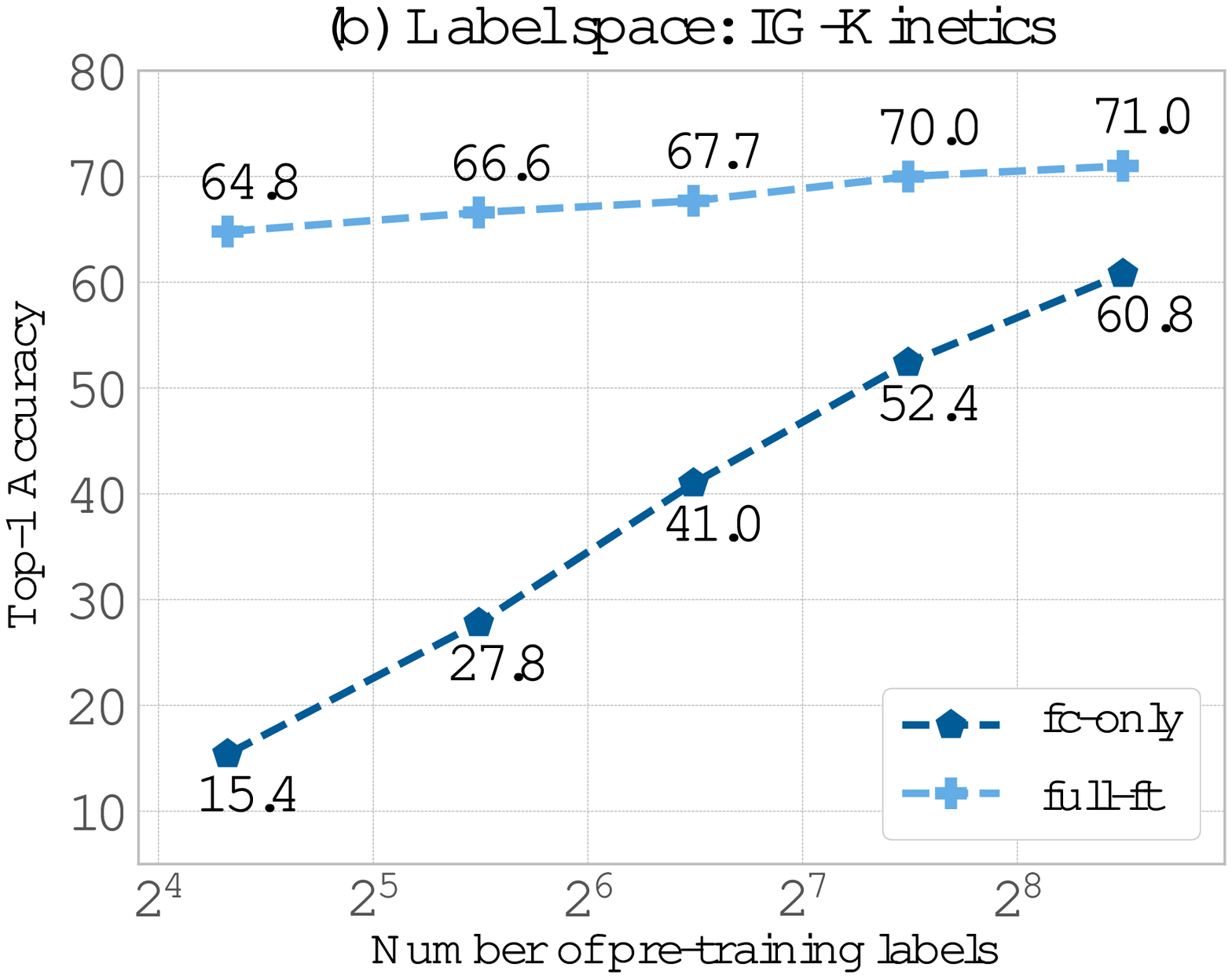} \\
\end{array}$ 
\caption{\scriptsize{Top-1 accuracy on Kinetics when pre-training on different number of labels. Note that the source datasets used in panels (a) and (b) are different, hence the results are not comparable. X-axis is log scale.}}
\vspace{-0.36in}
\label{fig:numLabels}
\end{center}
\end{figure}
In Sec. \ref{sec:datapoints}, we study how varying the number of pre-training videos for a fixed source label space effects the transfer learning performance. In this section, we investigate the reverse scenario, i.e., vary the number of pre-training labels while keeping the number of videos fixed. We consider \WSone{Verb+Noun} as our candidate pre-training dataset due to a large number ($10,653$) of labels. We randomly\footnote{Random sampling also makes sure that we remove uniformly from head and tail classes and long-tail issue with \WSone{Verb+Noun} does not affect the observations.} sub-sample different number of labels from the full label set all the way to $675$ labels and fix the number of videos in each resulting dataset to be $1M$. We did not have enough training videos, i.e., at least $1M$ for fewer than $675$ labels. Label sampling is done such that the smaller label space is a subset of the larger one. R(2+1)D-34 is used for pre-training with a clip-length of $8$.

Figure \ref{fig:numLabels} (a) shows performance on Kinetics. We may observe that using full-ft, there is an improvement of \texttildelow$1\%$ until $1350$ labels, following which the performance saturates. For fc-only approach, the improvement in accuracy is \texttildelow$9\%$ before it saturates at $2700$ labels. This suggests that the relatively fewer action labels in Kinetics ($400$) may not require a highly diverse and extensive pre-training label space such as \WSone{Verb+Noun}. However, a large image label space ($~17K$ hashtags) was proven~\cite{uruImage} to be effective for highly diverse target image tasks (e.g., ImageNet-5k). Hence, we believe that to reap the full benefits of a large pre-training video label space, there is a need for more diverse benchmark video datasets with large label space.

Next, to understand the effect when the number of pre-training labels are fewer than the target labels (i.e, \textless $400$ for Kinetics), we consider \WSone{Kinetics} as our pre-training dataset and vary the number of labels from $20$ to $360$. Pre-training data size is again fixed to $1M$. From Fig. \ref{fig:numLabels} (b), we may observe a log-linear behavior as we vary the number of labels. There is a significant drop in the performance when using fewer labels even in the full-ft evaluation setting. This indicates that pre-training on a small label space that is a subset of the target label space hampers performance.

In summary, while using fewer pre-training labels hurts performance (Fig. \ref{fig:numLabels} (b)), increasing the diversity through a simple approach of combining verbs and nouns (Fig. \ref{fig:numLabels} (a)) does not improve performance either. Thus, this analysis highlights the challenges in label space engineering, especially for video tasks.

\vspace{-0.05in}
\subsection{Exploring the temporal dimension of video} \label{sec:temporal}
We now explore the temporal aspects of videos over long and short time scales. As mentioned in Sec. \ref{sec:igdataset}, our dataset inherently has large amounts of \textit{temporal noise} as video lengths vary from $1$ -- $60$ seconds and no manual cleaning was undertaken. While short videos are better localized, longer videos can potentially contain more diverse content. First, we attempt to understand this trade-off between temporal noise and visual diversity. Second, we address a more fundamental question of whether video clip-based pre-training is needed at all or is frame-based pre-training followed by inflation~\cite{i3d} is sufficient. The latter has an advantage of being very fast and more scalable.
\vspace{-0.15in}
\subsubsection{Effect of temporal noise} \label{sec:vidLengths}
\vspace{-0.05in}
To study this, we construct $3$ datasets from \WSone{Kinetics}: \\
(i) \texttt{short-N}: $N$ videos of lengths between $1$ -- $5$ seconds.\\
(ii) \texttt{long-N}: $N$ videos of lengths between $55$ -- $60$ seconds.\\
(iii) \texttt{long-center-N}: $N$ videos ($4$ second long) constructed from the center portions of videos from \texttt{long-N}.\\
We ensure that the temporal dimension is the only factor that varies by keeping the label space and distribution (videos per label) fixed across these $3$ datasets. Temporal jittering is performed for all these datasets during pre-training. Also, note that the exact same number of videos are seen while training on all the datasets. We now consider the following two scenarios.\\
\noindent\textbf{Fixed number of videos budget (F1):} A natural question that arises is: given a fixed budget of videos, what temporal property should guide our selection of pre-training videos? To answer this, we fix the total number of unique videos to $5M$ and consider \texttt{short-5M}, \texttt{long-5M}, and \texttt{long-center-5M} datasets. Note that both \texttt{short-5M} and \texttt{long-center-5M} have similar per-video duration (i.e., $4$ seconds on average), but \texttt{long-center-5M} has greater temporal noise, since short videos are presumably more temporally localized than any given portion of longer videos. Between \texttt{short-5M} and \texttt{long-5M}, while \texttt{short-5M} has better temporal localization, \texttt{long-5M} may have greater content diversity 
From Table~\ref{tbl:vidLengths}, we may observe that \texttt{short-5M} performs significantly better than \texttt{long-center-5M} suggesting that short videos do provide better temporal localization. Also, \texttt{long-5M} performs better than \texttt{short-5M} by $3.2\%$ indicating that more diverse content in longer videos can mask the effect of temporal noise. Thus, for a fixed total number of videos, longer videos may benefit transfer learning than short videos.\vspace*{0.05in}\\
\noindent\textbf{Fixed video time budget (F2):} If storage or bandwidth is a concern, it is more practical to fix the total duration of videos, instead of the total number. Given this fixed budget of video hours, should we choose short or long videos? To answer this, we consider \texttt{short-5M}, \texttt{long-center-5M} and \texttt{long-500K} datasets, all with similar total video hours. From Table~\ref{tbl:vidLengths}, we observe that \texttt{short-5M} significantly outperforms \texttt{long-500K}. This indicates that diversity and/or temporal localization introduced by using more short videos is more beneficial than the diversity within fewer long videos. Thus, for a fixed video duration budget, choosing more short videos yields better results. \texttt{long-center-5M} and \texttt{long-500K} perform similarly, indicating that on average, a fixed central crop from a long video contains similar information to a random crop from a long video. \texttt{short-5M} outperforms \texttt{long-center-5M}, consistent with the claim that short videos do indeed have better temporal localization.
\begin{table}
\scriptsize
\setlength\extrarowheight{1.2pt}
\centering
\begin{center}
\begin{tabular}{|P{0.5cm}|P{1.1cm}|P{1.3cm}|P{1.2cm}|P{1.9cm}|}
\hline
& \texttt{long-5M} & \texttt{long-500K} & \texttt{short-5M} & \texttt{long-center-5M} \\\cline{1-3}
\hline
F1 & 60.6 & - & \multirow{ 2}{*}{57.4} & \multirow{ 2}{*}{51.4} \\\cline{1-3}
F2 & - & 50.6 & & \\
\hline
\end{tabular}
\end{center}
\vspace{-0.1in}
\caption{\scriptsize{Video top-1 accuracy when R(2+1)D-34 is pre-trained on $4$ different short and long video datasets, followed by fc-only finetuning on Kinetics.}}
\vspace{-0.24in}
\label{tbl:vidLengths}
\end{table}
\vspace{-0.15in}
\subsubsection{Frame- vs. clip-based pre-training:} \label{sec:image2video}
\vspace{-0.05in}
Although we have shown substantial gains when using clip-based R(2+1)D models for large-scale weakly supervised pre-training, it is computationally more intensive than $2D$ (image) models. Moreover, techniques such as \textit{inflation} \cite{i3d} efficiently leverage pre-trained image models by converting $2D$ filters to $3D$ and achieve top-performance on benchmark datasets. Given these, we want to understand the key value in pre-training directly on weakly-supervised video clips vs. images. 

Towards this end, we first construct an image variant of the \WSone{Kinetics} dataset (suffixed by $-Images$ in Table \ref{tbl:image2video1}), following the procedure described in Sec.~\ref{sec:igdataset}. We pre-train an $18$ layer $2D$ deep residual model ($R2D$) \cite{residual} from scratch on different types of $2D$ data (image/single video frames). We then \textit{inflate} \cite{i3d} this model to $R3D$\footnote{We chose to inflate to $R3D$ because it was not immediately obvious how to inflate a $2D$ model to R(2+1)D given that it factorizes $3D$ convolution to $2D$ spatial and $1D$ temporal \cite{r2plus1D}.} \cite{r2plus1D} and perform full-finetuning with a clip-length of $8$ on Kinetics. 

From the inflation-based models in Table \ref{tbl:image2video1}, we may observe that, pre-training on ImageNet achieves an improvement of $0.9\%$ compared to training $R3D$ from scratch, while pre-training on \WStwo{Kinetics}{19M}-Images yields a modest boost of $0.5\%$ over ImageNet. Training on random video frames from \WStwo{Kinetics}{19M} gives a further improvement of $0.5\%$ over weakly-supervised image pre-training and an overall boost of $1.0\%$ over ImageNet. To make sure that this marginal improvement is not because of pre-training on only $19M$ weakly-supervised noisy images, we pre-train using \WStwo{Kinetics}{250M}-Images but find no further improvements. Finally, pre-training $R3D$ directly using video clips achieves an accuracy of $71.7\%$, a significant jump of $4.2\%$ over the best inflated model ($67.5\%$). This clearly indicates that effectively modeling the temporal structure of videos in a very large-scale pre-training setup is extremely beneficial.
\begin{table}[t]
\scriptsize
\setlength\extrarowheight{1.2pt}
\centering
\begin{center}
\begin{tabular}{|P{2.63cm}|P{1.3cm}|P{0.88cm}|P{0.8cm}|P{0.6cm}|}
\hline
 Input dataset & Pre-training Input & Pre-train model & FT model & Top-1\\
\hhline{|=|=|=|=|=|}
ImageNet & Image & R2D-18 & R3D-18 & 66.5 \\
\WStwo{Kinetics}{19M}-Images & Image & R2D-18 & R3D-18 & 67.0 \\
\WStwo{Kinetics}{250M}-Images & Image & R2D-18 & R3D-18 & 67.0 \\
\WStwo{Kinetics}{19M} & Video frame & R2D-18 & R3D-18 & 67.5 \\
\hline
Kinetics & Video clip & R3D-18 & R3D-18 & 65.6 \\
\WStwo{Kinetics}{19M} & Video clip & R3D-18 & R3D-18 & \textbf{71.7} \\
\hline
\end{tabular}
\end{center}
\vspace{-0.09in}
\caption{\scriptsize{Understanding the benefit of using images vs. videos for pre-training.}}
\vspace{-0.24in}
\label{tbl:image2video1}
\end{table}
\vspace{-0.05in}
\subsection{Comparisons with state-of-the-art}\label{sec:sota}
In this section, we compare R(2+1)D-34 and R(2+1)D-152 models pre-trained on \WStwo{Kinetics}{65M} with several state-of-the-art approaches on $3$ different target datasets. For the results reported in this section alone,
we follow~\cite{i3d} to perform fully-convolutional prediction for a closely-fair comparison with other approaches. Specifically, the fully-connected layer in R(2+1)D is transformed into a $1 \times 1 \times 1$ convolutional layer (while retaining learned weights), to allow fully-convolutional evaluation. Each test video is scaled to $128 \times 171$, then cropped to $128 \times 128$ (a full center crop). We also report results from using another frame scaling approach (indicated as SE in Tables \ref{tbl:sotaKin} - \ref{tbl:sotasomething}), where each (train / test) video's shortest edge is scaled to $128$, while maintaining its original aspect ratio, followed by a full center crop. 

We note that each approach being compared varies greatly in terms of model architectures, pre-training datasets (ImageNet vs. Sports-1M), amount and type of input data (RGB vs flow vs audio, etc.), input clip size, input frame size, evaluation strategy, and so on. We also note that many prior state-of-the-art models use complex, optimized network architectures compared to ours. Despite these differences, our approach of pre-training on tens of millions of videos outperforms \underline{all} existing methods by a substantial margin of $\boldsymbol{3.6\%}$ when fully-finetuned on Kinetics (Table \ref{tbl:sotaKin}). Further, instead of uniformly sampling $10$ clips, we used SC-Sampler \cite{scsampler} and sampled 10 salient clips from test videos and achieved a top-1 accuracy of $\boldsymbol{82.8\%}$.

In Table \ref{tbl:sotaKitchens}, we report the performance on the validation~\cite{epic1}, seen (S1), and unseen (S2) test datasets that are part of the EPIC-Kitchens Action Recognition Challenge \cite{epicKitchens1}. Since the training dataset of EPIC-Kitchens consists of only \texttildelow$20K$ videos, for stronger baselines, we pre-train separate R(2+1)D-34 models on Kinetics and Sports-1M and fine-tune on EPIC-Kitchens. We also report the top-performing method from the challenge website \cite{epicKitchens1} at the time of this manuscript submission. From Table \ref{tbl:sotaKitchens}, we may observe that, on unseen kitchens (S2), R(2+1)D-152 pre-trained on \WStwo{Kinetics}{65M} improves the top-1 accuracy on verbs and nouns by $\boldsymbol{8.9\%}$ and $\boldsymbol{9.1\%}$ compared to R(2+1)D-34 pre-trained on Kinetics; and a $\boldsymbol{7.3\%}$ boost on actions compared to R(2+1)D-34 pre-trained on Sports-1M. Similar substantial gains hold for seen (S1) and validation datasets. We note that we process only $32$ RGB frames of the input video (no optical flow), at a much lower resolution ($128 \times 128$) compared to the state-of-the-art, which is an ensemble model.

Finally, we report the performance (Table~\ref{tbl:sotasomething}) on the validation data of Something-V1~\cite{something-something}, a challenging dataset with fine-grained classes. Using only RGB as input, pre-training with \WStwo{Kinetics}{65M} achieves a top-1 accuracy of $51.6\%$, an improvement of $\boldsymbol{2.1\%}$ over state-of-the-art \cite{somethingeco}\footnote{This number is achieved using RGB+flow and an ensemble of models.} ($49.5\%$). Compared to other approaches that use only RGB as input \cite{somethings3d}, our approach yields a boost of $3.4\%$. 
\begin{table}[t]
\scriptsize
\setlength\extrarowheight{1.2pt}
\centering
\begin{center}
\begin{tabular}{|P{4.2cm}|P{0.6cm}|P{0.6cm}|P{1.2cm}|}
\hline
Method; pre-training & top-1 & top-5 & Input type \\
\hhline{|=|=|=|=|}
I3D-Two-Stream \cite{i3d}; ImageNet & 75.7 & 92.0 & RGB + flow \\
R(2+1)D-Two-Stream~\cite{r2plus1D}; Sports-1M & 75.4 & 91.9 & RGB + flow \\
3-stream SATT \cite{kinetics17}; ImageNet & 77.7 & 93.2 & RGB + flow + audio \\
NL I3D \cite{nonlocal}; ImageNet & 77.7 & 93.3 & RGB \\
\hline
R(2+1)D-34; Sports-1M & 71.7 & 90.5 & RGB \\
Ours R(2+1)D-34; IG-Kinetics & 79.1 & 93.9 & RGB \\
{Ours R(2+1)D-34; IG-Kinetics; SE} & 79.6 & 94.2 & RGB \\
Ours R(2+1)D-152; IG-Kinetics & 80.5 & 94.6 & RGB \\
{Ours R(2+1)D-152; IG-Kinetics; SE} & 81.3 & 95.1 & RGB \\
{Ours R(2+1)D-152 + SC-Sampler \cite{scsampler}; IG-Kinetics; SE } & \textbf{82.8} & \textbf{95.3} & RGB \\
\hline
\end{tabular}
\end{center}
\vspace{-0.1in}
\caption{\scriptsize{Comparison with the state-of-the-art on Kinetics. SE: short edge scaling.}} \label{tbl:sotaKin}
\vspace{-0.25in}
\end{table}
\begin{table}[t]
\scriptsize
\setlength\extrarowheight{1.2pt}
\centering
\begin{center}
\begin{tabular}{|P{2.6cm}|P{0.44cm}|P{0.44cm}|P{0.44cm}|P{0.44cm}|P{0.44cm}|P{0.44cm}|P{0.44cm}|}
\hline
Method; pre-training & \multicolumn{2}{|c|}{Verbs} & \multicolumn{2}{|c|}{Nouns}  & \multicolumn{2}{|c|}{Actions}  \\
\hline
 &  \tiny top-1 & \tiny top-5 & \tiny top-1 & \tiny top-5 & \tiny top-1 & \tiny top-5\\
\hhline{|=|=|=|=|=|=|=|}
\multicolumn{7}{|c|}{\textit{Test Unseen (S2)}}  \\
\hline
Leaderboard~\cite{epicKitchens1} & 54.5 & \textbf{81.2} & 30.4 & 55.7 & 21.0 & 39.4 \\
\hline
\underline{R(2+1)D}~\cite{r2plus1D} & & & & & & \\
d=34; Kinetics & 48.4 & 77.2 & 26.6 & 50.4 & 16.8 & 31.2 \\
d=34; Sports-1M & 47.2 & 77.4 & 28.7 & 50.0 & 18.3 & 31.6 \\
Ours: d=34; IG-Kin. & 55.5 & 80.9 & 33.6 & 56.7 & 23.7 & 39.1 \\
{Ours: d=34; IG-Kin. ; SE} & 56.0 & 80.6 & 32.4 & 55.6 & 23.6 & 39.5  
\\
Ours: d=152; IG-Kin. & 55.3 & 80.3 & 34.7 & 58.2 & 25.4 & 40.7 \\
{Ours: d=152; IG-Kin.; SE} & \textbf{57.3} & 81.1 & \textbf{35.7} & \textbf{58.7} & \textbf{25.6} & \textbf{42.7}
\\
\hline
\multicolumn{7}{|c|}{\textit{Test Seen (S1)}}  \\
\hline
Leaderboard~\cite{epicKitchens1} & \textbf{66.1} & \textbf{91.3} & \textbf{47.9} & \textbf{72.8} & \textbf{36.7} & \textbf{58.6} \\
\hline
\underline{R(2+1)D}~\cite{r2plus1D} & & & & & & \\
d=34; Kinetics & 59.1 & 87.4 & 38.0 & 62.7 & 26.8 & 46.1 \\
d-34; Sports-1M & 59.6 & 87.2 & 43.7 & 67.0 & 31.0 & 50.3 \\
Ours: d=34; IG-Kin. & 63.3 & 87.5 & 46.3 & 69.6 & 34.4 & 54.2 \\
{Ours: d=34; IG-Kin. ; SE} & 63.2 & 87.6 & 45.4 & 68.7 & 33.4 & 52.4 \\
Ours: d=152; IG-Kin. & 63.8 & 87.7 & 45.3 & 68.3 & 34.1 & 53.5 \\
{Ours: d=152; IG-Kin.; SE} & 65.2 & 87.4 & 45.1 & 67.8 & 34.5 & 53.8 \\
\hline
\multicolumn{7}{|c|}{\textit{Validation}}  \\
\hline
Baradel \textit{et. al.} \cite{epic1}; - & 40.9 & - & - & - & - & - \\
\hline
\underline{R(2+1)D}~\cite{r2plus1D} & & & & & & \\
d=34; Kinetics & 46.8 & 79.2 & 25.6 & 47.5 & 15.3 & 29.4 \\
d=34; Sports-1M & 50.0 & 79.8 & 24.8 & 46.2 & 16.0 & 30.3 \\
Ours: d=34; IG-Kin. & 56.6 & 83.5 & 32.7 & 55.5 & 22.5 & 39.2 \\
{Ours: d=34; IG-Kin. ; SE} & 55.5 & 83.3 & 34.8 & 57.2 & 22.8 & 39.8 \\
Ours: d=152; IG-Kin. & 56.6 & 83.8 & 34.5 & 58.5 & 23.5 & 40.6 \\
{Ours: d=152; IG-Kin.; SE} & \textbf{58.4} & \textbf{84.1} & \textbf{36.9} & \textbf{60.3} & \textbf{26.1} & \textbf{42.7} \\
\hline
\end{tabular}
\end{center}
\vspace{-0.1in}
\caption{\scriptsize{Comparison with the state-of-the-art approaches on Epic-Kitchens dataset. IG-Kin. refers to IG-Kinetics. SE: short edge scaling.}} \label{tbl:sotaKitchens}
\vspace{-0.1in}
\end{table}
\begin{table}[t]
\scriptsize
\setlength\extrarowheight{1.2pt}
\centering
\begin{center}
\begin{tabular}{|P{3.5cm}|P{0.55cm}|P{0.55cm}|P{1.5cm}|}
\hline
Method; pre-training & top-1 & top-5 & Input type \\
\hhline{|=|=|=|=|}
NL I3D + Joint GCN \cite{somethinggvn} & 46.1 & 76.8 & RGB \\
S3D-G \cite{somethings3d} & 48.2 & 78.7 & RGB\\
ECO\textsubscript{En}Lite \cite{somethingeco} & 46.4 & - & RGB \\
ECO\textsubscript{En}Lite \cite{somethingeco}& 49.5 & - & RGB + flow \\
\hline
R(2+1)D-34; Kinetics & 45.2 & 74.1 & RGB \\
R(2+1)D-34; Sports-1M & 45.7 & 74.5 & RGB \\
Ours: R(2+1)D-34; IG-Kin. & 49.7 & 77.5 & RGB \\
{Ours: R(2+1)D-34; IG-Kin.; SE} & 49.9 & 77.5 & RGB \\
Ours: R(2+1)D-152; IG-Kin. & 51.0 & \textbf{79.0} & RGB \\
{Ours: R(2+1)D-152; IG-Kin.; SE} & \textbf{51.6} & 78.8 & RGB \\
\hline
\end{tabular}
\end{center}
\vspace{-0.1in}
\caption{\scriptsize{Comparison with the state-of-the-art on Something-V1 \cite{something-something}. IG-Kin. refers to IG-Kinetics. SE: short edge scaling.}} \label{tbl:sotasomething}
\vspace{-0.25in}
\end{table}

\vspace{-0.1in}
\section{Discussion} \label{sec:discuss}
\vspace{-0.05in}
In this work, we explored the feasibility of large-scale, noisy, weakly-supervised pre-training with tens of million of videos. Despite the presence of significant noise in label space and temporal localization, our pre-trained models learn very strong feature representations. These models are able to significantly improve the state-of-the-art action recognition results on the popular Kinetics \cite{kinetics}, a recently introduced EPIC-Kitchens \cite{epicKitchens}, and Something-something \cite{something-something} datasets. All of our large-scale pre-trained models show significant gains over Kinetics and Sports-1M, the de facto pre-training datasets in the literature. Our ablation studies address many important questions related to scale, label space, and temporal dimension, while also raising other interesting questions.

Our study of label spaces found that sampling from the joint distribution of verb-noun pairs performs relatively poorly; this is presumably due to the skewed distribution and suggests that solving low-shot learning at scale is an important area of investment. Data augmentation can also help here: social media offers a nearly unlimited supply of unlabeled video, and non-parametric approaches like K-nearest neighbors could be employed to enhance sparse labels. Optimizing label space granularity in the face of data sparsity is another worth-while direction; this may require finding algorithmic ways to construct hashtag taxonomies for videos. Future research should also invest in creating new public benchmarks with larger and richer label spaces. Label spaces of current target datasets are small; they do not reflect the value of large-scale pre-training. 

Finally, our analysis of temporal localization raises more questions. Our experiments clearly show the competing benefits of both good temporal localization in short videos, and greater diversity in long videos. Understanding this trade-off more rigorously may lead to intelligent data construction strategies that can leverage the best of both worlds.

{\small
\bibliographystyle{ieee}
\bibliography{egbib}

\begin{thebibliography}{10}\itemsep=-1pt

\bibitem{epicKitchens1}
{EPIC-Kitchens Action Recognition Challenge}.
\newblock [Online] Available
  \url{https://competitions.codalab.org/competitions/20115\#results}.

\bibitem{youtube8m}
S.~Abu-El-Haija, N.~Kothari, J.~Lee, P.~Natsev, G.~Toderici, B.~Varadarajan,
  and S.~Vijayanarasimhan.
\newblock Youtube-8m: A large-scale video classification benchmark.
\newblock {\em arXiv preprint arXiv:1609.08675}, 2016.

\bibitem{posetrack}
M.~Andriluka, U.~Iqbal, A.~Milan, E.~Insafutdinov, L.~Pishchulin, J.~Gall, and
  B.~Schiele.
\newblock Posetrack: A benchmark for human pose estimation and tracking.
\newblock In {\em CVPR}, 2018.

\bibitem{scsampler}
{B. Korbar, D. Tran, L. Torresani}.
\newblock {SCSampler}: Sampling salient clips from video for efficient action
  recognition.
\newblock {\em arXiv preprint arXiv:1904.04289}, 2019.

\bibitem{yfcc100m}
{B. Thomee, D. A. Shamma, G. Friedland, B. Elizalde, K. Ni, D. Poland, D.
  Borth, and L. Li}.
\newblock The new data and new challenges in multimedia research.
\newblock {\em CoRR, abs/1503.01817}, 2015.

\bibitem{epic1}
F.~Baradel, N.~Neverova, C.~Wolf, J.~Mille, and G.~Mori.
\newblock Object level visual reasoning in videos.
\newblock {\em arXiv preprint arXiv:1806.06157}, 2018.

\bibitem{bergamo2010exploiting}
A.~Bergamo and L.~Torresani.
\newblock Exploiting weakly-labeled web images to improve object
  classification: a domain adaptation approach.
\newblock In {\em NIPS}, 2010.

\bibitem{deepDetect}
H.~Bilen and A.~Vedaldi.
\newblock Weakly supervised deep detection networks.
\newblock In {\em CVPR}, 2016.

\bibitem{v3}
P.~Bojanowski, F.~Bach, I.~Laptev, J.~Ponce, C.~Schmid, and J.~Sivic.
\newblock Finding actors and actions in movies.
\newblock In {\em ICCV}, 2013.

\bibitem{caffe2}
{{Caffe2 Team}}.
\newblock {Caffe2 : A new lightweight, modular, and scalable deep learning
  framework}.
\newblock [Online] Available \url{https://caffe2.ai/}.

\bibitem{poseEstimation}
Z.~Cao, T.~Simon, S.-E. Wei, and Y.~Sheikh.
\newblock Realtime multi-person 2d pose estimation using part affinity fields.
\newblock {\em arXiv preprint arXiv:1611.08050}, 2016.

\bibitem{i3d}
J.~Carreira and A.~Zisserman.
\newblock Quo vadis, action recognition? a new model and the kinetics dataset.
\newblock {\em CVPR}, 2017.

\bibitem{chen2015webly}
X.~Chen and A.~Gupta.
\newblock Webly supervised learning of convolutional networks.
\newblock In {\em ICCV}, 2015.

\bibitem{neil}
X.~Chen, A.~Shrivastava, and A.~Gupta.
\newblock Neil: Extracting visual knowledge from web data.
\newblock In {\em ICCV}, 2013.

\bibitem{r2plus1D}
{D. Tran, H. Wang, L. Torresani, J. Ray, Y. LeCun, and M. Paluri}.
\newblock A closer look at spatiotemporal convolutions for action recognition.
\newblock {\em CVPR}, 2018.

\bibitem{segment1}
J.~Dai, K.~He, and J.~Sun.
\newblock Boxsup: Exploiting bounding boxes to supervise convolutional networks
  for semantic segmentation.
\newblock In {\em ICCV}, 2015.

\bibitem{epicKitchens}
D.~Damen, H.~Doughty, G.~M. Farinella, S.~Fidler, A.~Furnari, E.~Kazakos,
  D.~Moltisanti, J.~Munro, T.~Perrett, W.~Price, and M.~Wray.
\newblock Scaling egocentric vision: The epic-kitchens dataset.
\newblock In {\em ECCV}, 2018.

\bibitem{imagenet}
J.~Deng, W.~Dong, R.~Socher, L.-J. Li, K.~Li, and L.~Fei-Fei.
\newblock Imagenet: A large-scale hierarchical image database.
\newblock {\em CVPR}, 2009.

\bibitem{everything}
S.~K. Divvala, A.~Farhadi, and C.~Guestrin.
\newblock Learning everything about anything: Webly-supervised visual concept
  learning.
\newblock In {\em CVPR}, 2014.

\bibitem{paris}
C.~Doersch, S.~Singh, A.~Gupta, J.~Sivic, and A.~Efros.
\newblock What makes paris look like paris?
\newblock {\em ACM Transactions on Graphics}, 2012.

\bibitem{decaf}
J.~Donahue, Y.~Jia, O.~Vinyals, J.~Hoffman, N.~Zhang, E.~Tzeng, and T.~Darrell.
\newblock Decaf: A deep convolutional activation feature for generic visual
  recognition.
\newblock {\em ICML}, 2014.

\bibitem{v4}
O.~Duchenne, I.~Laptev, J.~Sivic, F.~Bach, and J.~Ponce.
\newblock Automatic annotation of human actions in video.
\newblock In {\em CVPR}. IEEE, 2009.

\bibitem{wildcat}
T.~Durand, T.~Mordan, N.~Thome, and M.~Cord.
\newblock Wildcat: Weakly supervised learning of deep convnets for image
  classification, pointwise localization and segmentation.
\newblock In {\em CVPR}, 2017.

\bibitem{v1}
M.~Everingham, J.~Sivic, and A.~Zisserman.
\newblock Hello! my name is buffy--automatic naming of characters in tv video.
\newblock 2006.

\bibitem{farhadi2010every}
A.~Farhadi, M.~Hejrati, M.~A. Sadeghi, P.~Young, C.~Rashtchian, J.~Hockenmaier,
  and D.~Forsyth.
\newblock Every picture tells a story: Generating sentences from images.
\newblock In {\em ECCV}, 2010.

\bibitem{fergus2010learning}
R.~Fergus, L.~Fei-Fei, P.~Perona, and A.~Zisserman.
\newblock Learning object categories from internet image searches.
\newblock {\em Proceedings of the IEEE}, 2010.

\bibitem{segment}
R.~Girshick, J.~Donahue, T.~Darrell, and J.~Malik.
\newblock Rich feature hierarchies for accurate object detection and semantic
  segmentation.
\newblock In {\em CVPR}, 2014.

\bibitem{imageNethour}
P.~Goyal, P.~Doll{\'a}r, R.~Girshick, P.~Noordhuis, L.~Wesolowski, A.~Kyrola,
  A.~Tulloch, Y.~Jia, and K.~He.
\newblock Accurate, large minibatch sgd: training imagenet in 1 hour.
\newblock {\em arXiv preprint arXiv:1706.02677}, 2017.

\bibitem{something-something}
R.~Goyal, S.~E. Kahou, V.~Michalski, J.~Materzynska, S.~Westphal, H.~Kim,
  V.~Haenel, I.~Fruend, P.~Yianilos, M.~Mueller-Freitag, et~al.
\newblock The ``something something" video database for learning and evaluating
  visual common sense.
\newblock In {\em ICCV}, 2017.

\bibitem{residual}
K.~He, X.~Zhang, S.~Ren, and J.~Sun.
\newblock Deep residual learning for image recognition.
\newblock In {\em CVPR}, pages 770--778, 2016.

\bibitem{segment3}
S.~Hong, J.~Oh, H.~Lee, and B.~Han.
\newblock Learning transferrable knowledge for semantic segmentation with deep
  convolutional neural network.
\newblock In {\em CVPR}, 2016.

\bibitem{segment2}
S.~Hong, D.~Yeo, S.~Kwak, H.~Lee, and B.~Han.
\newblock Weakly supervised semantic segmentation using web-crawled videos.
\newblock {\em arXiv preprint arXiv:1701.00352}, 2017.

\bibitem{imagenetTransfer}
M.~Huh, P.~Agrawal, and A.~A. Efros.
\newblock What makes imagenet good for transfer learning?
\newblock {\em arXiv preprint arXiv:1608.08614}, 2016.

\bibitem{batchNorm}
S.~Ioffe and C.~Szegedy.
\newblock Batch normalization: Accelerating deep network training by reducing
  internal covariate shift.
\newblock {\em ICML}, 2015.

\bibitem{f1}
H.~Izadinia, B.~C. Russell, A.~Farhadi, M.~D. Hoffman, and A.~Hertzmann.
\newblock Deep classifiers from image tags in the wild.
\newblock In {\em Workshop on Community-Organized Multimodal Mining:
  Opportunities for Novel Solutions}, 2015.

\bibitem{thumos}
Y.~Jiang, J.~Liu, A.~R. Zamir, G.~Toderici, I.~Laptev, M.~Shah, and
  R.~Sukthankar.
\newblock {THUMOS challenge: Action recognition with a large number of
  classes}.
\newblock 2014.
\newblock http://crcv.ucf.edu/THUMOS14.

\bibitem{laurens}
A.~Joulin, L.~van~der Maaten, A.~Jabri, and N.~Vasilache.
\newblock Learning visual features from large weakly supervised data.
\newblock {\em ECCV}, 2016.

\bibitem{f2}
A.~Joulin, L.~van~der Maaten, A.~Jabri, and N.~Vasilache.
\newblock Learning visual features from large weakly supervised data.
\newblock In {\em ECCV}, 2016.

\bibitem{sports1m}
A.~Karpathy, G.~Toderici, S.~Shetty, T.~Leung, R.~Sukthankar, and L.~Fei-Fei.
\newblock Large-scale video classification with convolutional neural networks.
\newblock {\em CVPR}, 2014.

\bibitem{segment4}
A.~Kolesnikov and C.~H. Lampert.
\newblock Seed, expand and constrain: Three principles for weakly-supervised
  image segmentation.
\newblock In {\em ECCV}, 2016.

\bibitem{signLang}
O.~Koller, H.~Ney, and R.~Bowden.
\newblock Deep hand: How to train a cnn on 1 million hand images when your data
  is continuous and weakly labelled.
\newblock In {\em CVPR}, 2016.

\bibitem{hmdb}
H.~Kuehne, H.~Jhuang, E.~Garrote, T.~Poggio, and T.~Serre.
\newblock {HMDB}: a large video database for human motion recognition.
\newblock {\em ICCV}, 2011.

\bibitem{kth}
I.~Laptev and T.~Lindeberg.
\newblock Space-time interest points.
\newblock {\em ICCV}, 2003.

\bibitem{hollywood}
I.~Laptev, M.~Marszalek, C.~Schmid, and B.~Rozenfeld.
\newblock Learning realistic human actions from movies.
\newblock {\em CVPR}, 2008.

\bibitem{f3}
A.~Li, A.~Jabri, A.~Joulin, and L.~van~der Maaten.
\newblock Learning visual n-grams from web data.
\newblock In {\em ICCV}, 2017.

\bibitem{l2}
Z.~Li and D.~Hoiem.
\newblock Learning without forgetting.
\newblock {\em PAMI}, 2017.

\bibitem{uruImage}
D.~Mahajan, R.~Girshick, V.~Ramanathan, K.~He, M.~Paluri, Y.~Li, A.~Bharambe,
  and L.~van~der Maaten.
\newblock Exploring the limits of weakly supervised pretraining.
\newblock {\em ECCV}, 2018.

\bibitem{v2}
M.~Marszalek, I.~Laptev, and C.~Schmid.
\newblock Actions in context.
\newblock In {\em CVPR}, 2009.

\bibitem{sqrtSampling}
T.~Mikolov, I.~Sutskever, K.~Chen, G.~S. Corrado, and J.~Dean.
\newblock Distributed representations of words and phrases and their
  compositionality.
\newblock In {\em NIPS}, 2013.

\bibitem{oflow}
J.~Y.-H. Ng, J.~Choi, J.~Neumann, and L.~S. Davis.
\newblock Actionflownet: Learning motion representation for action recognition.
\newblock In {\em WACV}, 2018.

\bibitem{freeLocalization}
M.~Oquab, L.~Bottou, I.~Laptev, and J.~Sivic.
\newblock Is object localization for free?-weakly-supervised learning with
  convolutional neural networks.
\newblock In {\em CVPR}, 2015.

\bibitem{im2text}
V.~Ordonez, G.~Kulkarni, and T.~L. Berg.
\newblock Im2text: Describing images using 1 million captioned photographs.
\newblock In {\em NIPS}, 2011.

\bibitem{segment5}
D.~Pathak, P.~Krahenbuhl, and T.~Darrell.
\newblock Constrained convolutional neural networks for weakly supervised
  segmentation.
\newblock In {\em ICCV}, 2015.

\bibitem{segment6}
D.~Pathak, E.~Shelhamer, J.~Long, and T.~Darrell.
\newblock Fully convolutional multi-class multiple instance learning.
\newblock {\em arXiv preprint arXiv:1412.7144}, 2014.

\bibitem{visualRelations}
J.~Peyre, I.~Laptev, C.~Schmid, and J.~Sivic.
\newblock Weakly-supervised learning of visual relations.
\newblock In {\em ICCV}, 2017.

\bibitem{segment7}
P.~O. Pinheiro and R.~Collobert.
\newblock From image-level to pixel-level labeling with convolutional networks.
\newblock In {\em CVPR}, 2015.

\bibitem{weakVideo}
A.~Prest, C.~Leistner, J.~Civera, C.~Schmid, and V.~Ferrari.
\newblock Learning object class detectors from weakly annotated video.
\newblock In {\em CVPR}, 2012.

\bibitem{noise1}
S.~Reed, H.~Lee, D.~Anguelov, C.~Szegedy, D.~Erhan, and A.~Rabinovich.
\newblock Training deep neural networks on noisy labels with bootstrapping.
\newblock {\em arXiv preprint arXiv:1412.6596}, 2014.

\bibitem{schroff2011harvesting}
F.~Schroff, A.~Criminisi, and A.~Zisserman.
\newblock Harvesting image databases from the web.
\newblock {\em PAMI}, 2011.

\bibitem{shi2017transfer}
Z.~Shi, P.~Siva, and T.~Xiang.
\newblock Transfer learning by ranking for weakly supervised object annotation.
\newblock {\em arXiv preprint arXiv:1705.00873}, 2017.

\bibitem{ucf}
K.~Soomro, A.~R. Zamir, and M.~Shah.
\newblock {UCF101}: A dataset of 101 human actions classes from videos in the
  wild.
\newblock {\em CRCV-TR-12-01}, 2012.

\bibitem{noise2}
S.~Sukhbaatar, J.~Bruna, M.~Paluri, L.~Bourdev, and R.~Fergus.
\newblock Training convolutional networks with noisy labels.
\newblock {\em arXiv preprint arXiv:1406.2080}, 2014.

\bibitem{jft}
C.~Sun, A.~Shrivastava, S.~Singh, and A.~Gupta.
\newblock Revisiting unreasonable effectiveness of data in deep learning era.
\newblock {\em ICCV}, 2017.

\bibitem{verbnet}
{{VerbNet}}.
\newblock {VerbNet : A Computational Lexical Resource for Verbs}.
\newblock [Online] Available \url{https://verbs.colorado.edu/verbnet/}.

\bibitem{kinetics}
{W. Kay, J. Carreira, K. Simonyan, B. Zhang, C. Hillier, S. Vijayanarasimhan,
  F. Viola, T. Green, T. Back, P. Natsev, M. Suleyman, and A. Zisserman}.
\newblock The kinetics human action video dataset.
\newblock {\em arXiv:1705.06950}, 2017.

\bibitem{lsh}
J.~Wang, H.~T. Shen, J.~Song, and J.~Ji.
\newblock Hashing for similarity search: A survey.
\newblock {\em arXiv preprint arXiv:1408.2927}, 2014.

\bibitem{nonlocal}
X.~Wang, R.~Girshick, A.~Gupta, and K.~He.
\newblock Non-local neural networks.
\newblock {\em CVPR}, 2018.

\bibitem{somethinggvn}
X.~Wang and A.~Gupta.
\newblock Videos as space-time region graphs.
\newblock {\em arXiv preprint arXiv:1806.01810}, 2018.

\bibitem{annotate}
X.-J. Wang, L.~Zhang, X.~Li, and W.-Y. Ma.
\newblock Annotating images by mining image search results.
\newblock {\em PAMI}, 2008.

\bibitem{somethings3d}
S.~Xie, C.~Sun, J.~Huang, Z.~Tu, and K.~Murphy.
\newblock Rethinking spatiotemporal feature learning for video understanding.
\newblock {\em arXiv preprint arXiv:1712.04851}, 2017.

\bibitem{kinetics17}
{Y. Bian, C. Gan, X. Liu, F. Li, X. Long, Y. Li, H. Qi, J. Zhou, S. Wen, and Y.
  Lin}.
\newblock Revisiting the effectiveness of off-the-shelf temporal modeling
  approaches for large-scale video classification.
\newblock {\em arXiv:1708.03805, 2017}, 2017.

\bibitem{census}
R.~Zabih and J.~Woodfill.
\newblock Non-parametric local transforms for computing visual correspondence.
\newblock In {\em ECCV}, 1994.

\bibitem{somethingeco}
M.~Zolfaghari, K.~S. Singh, and T.~Brox.
\newblock {ECO:} efficient convolutional network for online video
  understanding.
\newblock {\em arXiv preprint arXiv:1804.09066}, 2018.

\end{thebibliography}
}
\section*{A Supplementary Material}
\subsection*{A.1 Dataset construction}
$\boldsymbol{\mathtt{IG-Verb-Noun}}$ dataset: When constructing this dataset, we combine the canonicalized forms of verbs and nouns and consider this as the class label. We construct the set $\mathtt{relevant\_hashtags}$ by considering all possible forms (with and without canonicalization) of the verbs and nouns. For example, for the verb `burning' and noun `candle,' $\mathtt{relevant\_hashtags} = \{ burncandle, candlesburning, candleburning, \\
burncandles, burningcandle, burningcandles, \\ candlesburn, candleburn \}$.
\subsection*{A.2 Video deduplication}
We devise a pipeline to deduplicate the source content that may overlap with any of the target dataset videos. Towards this end, we first decode videos at $16$ fps and scale to a resolution of $112 \times 112$. Next, we employ the Census Transform technique \cite{census} to extract a $64$ dimensional per-frame global feature vector for each videos across all source and target datasets. Following this, we use locality sensitive hashing \cite{lsh} to measure the amount of overlap (percentage of feature match w.r.t. source video length) between each source and target video. Census Transform is robust to scale thereby allowing us to identify scenarios where the target video is a scaled version of a source video. Furthermore, operating on features at a frame-level also help tackle scenarios where a small portion of the target video may occur anywhere in a source video.

For each video from the three target datasets, we compute a percentage of overlap with each video in the source dataset. Using this approach, we identify about $20K$ videos from \WStwo{Kinetics}{65M} dataset that potentially overlap (with varying amounts) with the three target datasets we consider: Kinetics, EPIC-Kitchens, and Something-something-v1. Upon manual inspection, some of videos from the source dataset that overlap with a high match percentage are truly duplicates. Nevertheless, to err on the side of caution, we adopt an aggressive low-precision high-recall strategy and exclude all $20K$ videos from \WStwo{Kinetics}{65M} dataset.
\\
\subsection*{A.3 Model architectural details}
We note that there is a slight variation in the architecture between models of depths $18$ and $34$ when compared to models of depth $> 34$. Specifically, for computation feasibility, as described in \cite{residual}, we use bottleneck blocks consisting of $1 \times 1 \times 1$ convolution before and after the factorized convolutions for R(2+1)D-101 and R(2+1)D-152. \subsection*{A.4 Hyper parameters}
Here, we report the hyper parameters we use for all our full-finetuning experiments. The total training length is the number of times we iterate over an entire target dataset through the network. To set the learning rate (LR), we follow the linear scaling rule with gradual warmup described in \cite{imageNethour} when the target dataset is Kinetics. We did not use warmup for EPIC-Kitchens and Something-something-v1. In a distributed environment, the final LR is scaled according to the total number of nodes and the number of GPUs per node ($16$ and $4$ respectively in our experiments). The LR is also decayed from the initial value by the specified factor at equally spaced steps; the total number LR decay steps is given in the LR steps column. We use a weight decay of $0.0001$ and LR decay of $0.1$ for all experiments. We report the hyper parameters when target dataset is Kinetics in Tables \ref{tbl:datapointsKinetics} -- \ref{tbl:image2video}, EPIC-Kitchens in Tables \ref{tbl:datapointsKitchens} -- \ref{tbl:sotaKitchens}, and Something-something-v1 in Table \ref{tbl:sotaSomething}. Since we train two separate models for the verb and noun tasks of EPIC-Kitchens, we report the LR used for them separately.
\begin{table}
\scriptsize
\setlength\extrarowheight{1.0pt}
\centering
\begin{center}
\begin{tabular}{|P{3.3cm}|P{0.45cm}|P{0.7cm}|P{0.68cm}|P{0.6cm}|}
\hline
Source dataset & Total length & LR per GPU & warmup & LR steps \\
\hline
IG-Kinetics--\{500K, 1M, 5M\} & 58 & 0.0003 & 16 & [16,16] \\
IG-Kinetics--\{10M, 19M, 65M\} & 58 & 0.00003 & 16 & [16,16] \\
\hline
\end{tabular}
\end{center}
\vspace{-0.1in}
\caption{\scriptsize{Hyper parameters for the ablation study on studying the\textit{ effect of amount of pre-training data (Sec. 4.1.1)} on transfer learning performance. Target dataset: Kinetics; Model: R(2+1)D-34; input video clip-length=8. batch size per GPU: 16. Data sampling strategy: Either random or tail-preserving.}} \label{tbl:datapointsKinetics}
\end{table}
\begin{table}

\scriptsize
\setlength\extrarowheight{1.0pt}
\centering
\begin{center}
\begin{tabular}{|P{2.8cm}|P{0.45cm}|P{0.7cm}|P{0.68cm}|P{0.6cm}|}
\hline
Model & Total length & LR per GPU & warmup & LR steps \\
\hline
R(2+1)D--\{18,34,101,152\} & 58 & 0.00003 & 16 & [16,16] \\
\hline
\end{tabular}
\end{center}
\vspace{-0.1in}
\caption{\scriptsize{Hyper parameters for the ablation study on studying the \textit{effect of model capacity (Sec. 4.1.2)} on transfer learning performance. Target dataset: Kinetics; Pre-training data: \WStwo{Kinetics}{65M}. Model: R(2+1)D; input video clip-length=32. batch size per GPU: 6.}} \label{tbl:depthKinetics}
\end{table}

\begin{table}
\scriptsize
\setlength\extrarowheight{1.0pt}
\centering
\begin{center}
\begin{tabular}{|P{3.2cm}|P{0.45cm}|P{0.7cm}|P{0.68cm}|P{0.6cm}|}
\hline
Number of labels & Total length & LR per GPU & warmup & LR steps \\
\hline
\{675,1350, 2700, 5400, 10563\} & 58 & 0.0003 & 16 & [16,16] \\
\hline
\end{tabular}
\end{center}
\vspace{-0.1in}
\caption{\scriptsize{Hyper parameters for the ablation study on studying the \textit{effect of the number of pre-training labels (Sec. 4.2.2)} on transfer learning performance. Source label space: \WSone{Verb+Noun}.
Target dataset: Kinetics; Model: R(2+1)D-34; input video clip-length=8. batch size per GPU: 16.}} \label{tbl:labels1}
\end{table}
\begin{table}
\scriptsize
\setlength\extrarowheight{1.0pt}
\centering
\begin{center}
\begin{tabular}{|P{3.2cm}|P{0.45cm}|P{0.7cm}|P{0.68cm}|P{0.6cm}|}
\hline
Number of labels & Total length & LR per GPU & warmup & LR steps \\
\hline
\{20, 45, 90, 180, 359\} & 58 & 0.0003 & 16 & [16,16] \\
\hline
\end{tabular}
\end{center}
\vspace{-0.1in}
\caption{\scriptsize{Hyper parameters for the ablation study on studying the \textit{effect of the number of pre-training labels (Sec. 4.2.2)} on transfer learning performance. Source label space: \WSone{Kinetics}.
Target dataset: Kinetics; Model: R(2+1)D-34; input video clip-length=8. batch size per GPU: 16.}} \label{tbl:labels2}
\end{table}
\begin{table}
\scriptsize
\setlength\extrarowheight{1.0pt}
\centering
\begin{center}
\begin{tabular}{|P{3cm}|P{0.45cm}|P{0.7cm}|P{0.65cm}|P{0.9cm}|}
\hline
Source dataset & Total length & LR per GPU & warmup & LR steps \\
\hline
ImageNet & 167 & 0.01 & 20 & [40,40,40] \\
\hline
\WStwo{Kinetics}{19M}-Images & 96 & 0.00025 & 20 & [20,20,20] \\
\hline
\WStwo{Kinetics}{250M}-Images & 96 & 0.0025 & 20 & [20,20,20] \\
\hline
\WStwo{Kinetics}{19M} (VideoFrames) & 96 & 0.0005 & 20 & [20,20,20] \\
\hhline{|=|=|=|=|=|}
\WStwo{Kinetics}{19M} (Video) & 188 & 0.01 & 40 & [40,40,40] \\
\hline
\end{tabular}
\end{center}
\vspace{-0.1in}
\caption{\scriptsize{Hyper parameters for the ablation study on studying \textit{frame vs. clip-based pre-training (Sec. 4.3.2)} on transfer learning performance. Target dataset: Kinetics}} \label{tbl:image2video}

\end{table}
\begin{table}
\scriptsize
\setlength\extrarowheight{1.0pt}
\centering
\begin{center}
\begin{tabular}{|P{3cm}|P{0.45cm}|P{0.69cm}|P{0.62cm}|P{0.54cm}|}
\hline
Method; Pre-training & Total length & LR per GPU & warmup & LR steps \\
\hline
R(2+1)D-34; Sports-1M & 58 & 0.00003 & 16 & [16,16] \\
\hline
R(2+1)D-34; IG-Kinetics & 58 & 0.00003 & 16 & [16,16] \\
\hline
R(2+1)D-152; IG-Kinetics & 58 & 0.00003 & 16 & [16,16] \\
\hline
\end{tabular}
\end{center}
\vspace{-0.1in}
\caption{\scriptsize{Hyper parameters to \textit{compare with the state-of-the-art methods (reported in Sec. 4.4)}. Target dataset: Kinetics; Input video clip-length=32. batch size per GPU: 6.}} \label{tbl:sotaKinetics}
\end{table}
\begin{table}
\scriptsize
\setlength\extrarowheight{1.0pt}
\centering
\begin{center}
\begin{tabular}{|P{1.8cm}|P{0.45cm}|P{1.5cm}|P{0.68cm}|P{0.6cm}|}
\hline
Source dataset & Total length & LR per GPU (Verb / Noun) & LR steps \\
\hline
\WStwo{Kinetics}{500K} & 27 & 0.0005/0.0005 & [9,9,9] \\
\WStwo{Kinetics}{1M} & 27 & 0.0005/0.0005 & [9,9,9] \\
\WStwo{Kinetics}{5M} & 27 & 0.0005/0.0001 & [9,9,9] \\
\WStwo{Kinetics}{10M} & 27 & 0.00025/0.0001 & [9,9,9] \\
\WStwo{Kinetics}{19M} & 27 & 0.00025/0.0001 & [9,9,9] \\
\WStwo{Kinetics}{65M} & 27 & 0.00025/0.0001 & [9,9,9] \\
Kinetics & 27 & 0.00025/0.0001 & [9,9,9] \\
\hline
\end{tabular}
\end{center}
\vspace{-0.1in}
\caption{\scriptsize{Hyper parameters for the ablation study on studying the\textit{ effect of amount of pre-training data (Sec. 4.1.1)} on transfer learning performance. Target dataset: EPIC-Kitchens; Model: R(2+1)D-34; input video clip-length=8. batch size per GPU: 16. Data sampling strategy: Either random or tail-preserving.}} \label{tbl:datapointsKitchens}
\end{table}

\begin{table}
\scriptsize
\setlength\extrarowheight{1.0pt}
\centering
\begin{center}
\begin{tabular}{|P{1.8cm}|P{0.45cm}|P{1.8cm}|P{0.6cm}|}
\hline
Model & Total length & LR per GPU & LR steps \\
\hline
R(2+1)D-18 & 27 & 0.00005/0.00025 & [9,9,9] \\
R(2+1)D-34 & 27 & 0.000025/0.00025 & [9,9,9] \\
R(2+1)D-101 & 27 & 0.000025/0.0001 & [9,9,9] \\
R(2+1)D-152 & 27 & 0.00005/0.0001 & [9,9,9] \\
\hline
\end{tabular}
\end{center}
\vspace{-0.1in}
\caption{\scriptsize{Hyper parameters for the ablation study on studying the \textit{effect of model capacity (Sec. 4.1.2)} on transfer learning performance. Target dataset: EPIC-Kitchens; Pre-training data: \WStwo{Kinetics}{65M}. Model: R(2+1)D; input video clip-length=32. batch size per GPU: 6.}} \label{tbl:depthKitchens}
\end{table}
\begin{table}
\scriptsize
\setlength\extrarowheight{1.0pt}
\centering
\begin{center}
\begin{tabular}{|P{3cm}|P{0.45cm}|P{1.8cm}|P{0.54cm}|}
\hline
Method; Pre-training & Total length & LR per GPU (Verb /Noun) & LR steps \\
\hline
R(2+1)D-34; Kinetics & 27 & 0.0025/0.0025 & [9,9,9] \\
\hline
R(2+1)D-34; Sports-1M & 27 & 0.0025/0.0025 & [9,9,9] \\
\hline
R(2+1)D-34; IG-Kinetics & 27 & 0.000025/0.00025 & [9,9,9] \\
\hline
R(2+1)D-152; IG-Kinetics & 27 & 0.00005/0.0001 & [9,9,9] \\
\hline
\end{tabular}
\end{center}
\vspace{-0.1in}
\caption{\scriptsize{Hyper parameters to \textit{compare with the state-of-the-art methods (reported in Sec. 4.4)}. Target dataset: EPIC-Kitchens; Input video clip-length=32. batch size per GPU: 6.}} \label{tbl:sotaKitchens}
\end{table}

\begin{table}[h]
\scriptsize
\setlength\extrarowheight{1.0pt}
\centering
\begin{center}
\begin{tabular}{|P{2.8cm}|P{0.45cm}|P{0.69cm}|P{0.54cm}|}
\hline
Method; Pre-training & Total length & LR per GPU & LR steps \\
\hline
R(2+1)D-34; Kinetics & 16 & 0.001 & [6,6] \\
R(2+1)D-34; Sports-1M & 16 & 0.001 & [6,6] \\
R(2+1)D-34; IG-Kinetics & 16 & 0.00025 & [6,6] \\
R(2+1)D-152; IG-Kinetics & 16 & 0.00013 & [12] \\
\hline
\end{tabular}
\end{center}
\vspace{-0.1in}
\caption{\scriptsize{Hyper parameters to \textit{compare with the state-of-the-art methods (reported in Sec. 4.4)}. Target dataset: Something-something-v1; Input video clip-length=32. batch size per GPU: 6.}} \label{tbl:sotaSomething}
\end{table}

\end{document}